\documentclass{article}

\usepackage{microtype}
\usepackage{graphicx}
\usepackage{subfigure}
\usepackage{booktabs} %
\usepackage{enumitem}

\usepackage{amsmath,amsfonts,bm}

\def\eqref#1{equation~\ref{#1}}

\def\1{\bm{1}}

\DeclareMathAlphabet{\mathsfit}{\encodingdefault}{\sfdefault}{m}{sl}
\SetMathAlphabet{\mathsfit}{bold}{\encodingdefault}{\sfdefault}{bx}{n}

\newcommand{\E}{\mathbb{E}}

\DeclareMathOperator*{\argmin}{arg\,min}

\usepackage{hyperref}

\usepackage[accepted]{icml2024}

\usepackage{amsmath}
\usepackage{amssymb}
\usepackage{mathtools}
\usepackage{amsthm}

\usepackage{thm-restate}

\usepackage[capitalize,noabbrev]{cleveref}

\theoremstyle{plain}
\newtheorem{theorem}{Theorem}[section]

\theoremstyle{definition}
\newtheorem{definition}[theorem]{Definition}

\theoremstyle{remark}

\global\long\def\Jo{J^{\textup{O}}}%
\global\long\def\Js{J^{\textup{S}}}%

\usepackage[textsize=tiny]{todonotes}

\icmltitlerunning{Learning to Stabilize Online Reinforcement Learning in Unbounded State Spaces}

\usepackage{todonotes}
\usepackage[showdeletions]{color-edits}
\addauthor[Matthew]{mz}{red}
\addauthor[Yudong]{yc}{blue}
\addauthor[Josiah]{jh}{red}
\addauthor[Qiaomin]{qx}{red}
\addauthor[Brahma]{bp}{red}

\begin{document}

\twocolumn[

\icmltitle{Learning to Stabilize Online Reinforcement Learning in Unbounded State Spaces}

\begin{icmlauthorlist}
\icmlauthor{Brahma S. Pavse}{uw}
\icmlauthor{Matthew  Zurek}{uw}
\icmlauthor{Yudong Chen}{uw}
\icmlauthor{Qiaomin Xie}{uw}
\icmlauthor{Josiah P. Hanna}{uw}
\end{icmlauthorlist}

\icmlaffiliation{uw}{University of Wisconsin--Madison, USA}

\icmlcorrespondingauthor{Brahma S. Pavse}{pavse@wisc.edu}

\icmlkeywords{reinforcement learning, unbounded state spaces, stability, queueing systems, reward shaping, reset-free RL, average-reward RL, lyapunov function, potential function}

\vskip 0.3in
]

\printAffiliationsAndNotice{}  %

\begin{abstract}
In many reinforcement learning (\textsc{rl}) applications, we want policies that reach desired states and then keep the controlled system within an acceptable region around the desired states over an indefinite period of time.
This latter objective is called \textit{stability} and is especially important when the state space is unbounded, such that the states can be arbitrarily far from each other and the agent can drift far away from the desired states.
For example, in stochastic queuing networks, where queues of waiting jobs can grow without bound, the desired state is all-zero queue lengths. Here, a stable policy ensures queue lengths are finite while an optimal policy minimizes queue lengths.
Since an optimal policy is also stable, one would expect that \textsc{rl} algorithms would implicitly give us stable policies. However, in this work, we find that deep \textsc{rl} algorithms that directly minimize the distance to the desired state during online training often result in unstable policies, i.e., policies that drift far away from the desired state. We attribute this instability to poor credit-assignment for destabilizing actions. We then introduce an approach based on two ideas: 1) a Lyapunov-based cost-shaping technique and 2) state transformations to the unbounded state space. 
We conduct an empirical study on various queuing networks and traffic signal control problems and find that our approach performs competitively against strong baselines with knowledge of the transition dynamics. Our code is available here: \url{https://github.com/Badger-RL/STOP}.
\end{abstract}

\section{Introduction}
\label{sec:intro}

Much of the recent progress in reinforcement learning (\textsc{rl}) has focused on obtaining optimal performance on tasks that can be completed in some amount of time.
However, there are many real-world sequential decision-making problems in which we want the learning agent to operate optimally over an indefinite period of time.
For example, in traffic intersection management, we want a policy that reaches and stays at the desired state of minimum waiting time of cars over an indefinite period of time.
Such problems are also common in control of stochastic queue networks and industrial process control \citep{srikant_comm_2014, neely_networkcomm_2010}.
While it is important that the agent is optimal in these problems, it is simultaneously important that the learning agent is \textit{stable}, meaning that its decisions keep the system in an acceptable region of the state space near the desired state~\citep{meynCSRL2022}.

Throughout this paper, we use queuing as a motivating example for stochastic environments with unbounded state spaces. In stochastic queuing networks, a server policy must select from several queues to serve based on the current number of jobs waiting in each queue and the desired state is all-zero queue lengths.
Ideally, the server keeps the average length of queues short so that no job must wait a long time to be served.
Jobs arrive continuously and --- if the server makes poor decisions (e.g., serving an empty queue) --- then the average length of the queues can grow infinitely-long.
In this case, an optimal policy minimizes the average queue length over time, while a stable policy ensures the queue lengths remain finite. By implication, an optimal policy is also a stable policy.

Since optimal behavior implies stable behavior, we would expect that directly trying to be optimal would implicitly give us stable behavior. 
However, in this work, we find that agents that do so actually learn unstable policies on multiple real-world-inspired domains. In particular, we empirically show that online deep \textsc{rl} agents that minimize the optimality cost objective of the domain drift far away from the desired state. We then propose an approach that specifically encourages the agent to be stable and optimal. More concretely, our contributions are as follows:
\begin{enumerate}[topsep=0pt, itemsep=0pt, parsep=0pt]
    \item We demonstrate that solely minimizing the optimality cost objective inadequately discourages destabilizing actions, which leads to the agent taking those actions early in learning. This challenge is especially problematic in the: 1) online learning setting, where destabilizing actions must be discouraged as quickly as possible to prevent the system's state from destabilizing, and 2) unbounded state space setting where the neural network's inability to effectively generalize across arbitrarily far unbounded samples prevents accurate credit assignment~\citep{challot_deep_2017}.
    \item We introduce STability and OPtimality (\textsc{stop}), an approach that encourages stability and optimality. \textsc{stop} is based on the combination of two techniques: 1) a Lyapunov-inspired cost shaping approach that explicitly encourages the agent to be stable and optimal and 2) state transformations that compress the unbounded state space to mitigate the burden of extreme generalization on neural networks. The first technique discourages destabilizing actions more than the true optimality cost function, even without knowing which actions are destabilizing. We provide intuition on how to choose an appropriate Lyapunov function by drawing upon results from queueing theory. The second technique improves the agent's ability to generalize across the unbounded state-space.
    \item We then prove that our cost-shaping approach has the desired consistency property that it does not change the optimal policy. This result is analogous to the potential-based shaping result of \citet{ngpotential1999}, but our result applies to the average-reward, unbounded state space setting, and relates consistency to stability.
    \item We conduct a thorough empirical study and analyze the role of different components of \textsc{stop} on the challenging real-world-inspired domains of queuing and traffic intersection management. We show that \textsc{stop} enables learning of highly performant online \textsc{rl} policies and, in some cases, outperforms algorithms that have knowledge of the transition dynamics.
\end{enumerate}

\section{Related Work}
\label{sec:related}

\textbf{Online RL.}
Our work focuses on online (or continuing) \textsc{rl} tasks in which interaction never terminates and performance is measured online \citep{sutton_rlbook_2018}.
This setting has been described with the autonomous \textsc{rl} \citep{sharma_autorl_2021} or single-life \textsc{rl} \citep{chen_you_2022} formalisms.
In many practical set-ups, it is infeasible to reset the agent to a new initial state. For example, manually placing a robot in an initial state or removing all vehicles at a traffic intersection is costly or even inappropriate. Recent work has thus considered \textit{reset-free} \textsc{rl} where the agent learns to reset itself \citep{eysenbach_leave_2018,zhu_ingredients_2020,gupta_reset-free_2021,han_learning_2015, sharma2022statedistribution}. However, these works still require a manual reset \citep{eysenbach_leave_2018}, use policies learned on one task as a reset policy in another task \citep{gupta_reset-free_2021}, or require access to  demonstrations from a target policy~\citep{sharma2022statedistribution}. Our work performs no resets and evaluates performance on a single infinitely-long, stationary task.

In online \textsc{rl}, the natural performance measure of an agent is average-reward \citep{sutton_rlbook_2018,naik_discnotopt_2019}.  Our work is different from prior average-reward \textsc{rl}  work \citep{mahadevan_average_1996,schwartz_reinforcement_1993, wan_learning_2021,wei_model-free_2020,zhang_artrpo_2021,zhang_average-reward_2021} in that we focus on the challenge of learning stable behavior in an unbounded state space.

\textbf{Stability and Unbounded State Spaces in RL.}
Stability is related to the notion of \emph{safety} in \textsc{rl}~\citep{garcia2015comprehensive}, but with crucial differences. Safety is typically defined as hard constraints on individual states and/or actions ~\citep{hans2008safe,dalal2018safe,dean2019safely,koller2018learning}, or soft constraints on the expected costs over the trajectory~\citep{achiam2017constrained,chow2018lyapunov,yu2019convergent}. In contrast, stability concerns the long-run asymptotic behavior of the system, which cannot be immediately written as constraints over the states and actions or a budget for some cost. 

Some work considers control-theoretic notions of stability~\citep{vinogradska2016stability,berkenkamp2017safe}. While related, these results mostly consider systems with deterministic and partially unknown dynamics. For example, \citet{westenbroek_lyapunov_2022} propose a cost-shaping approach with similarities to ours, but focusing on deterministic dynamics; they consider discounted costs, and do not study the continuing setting with unbounded states.

The work of \citet{shah_unbounded_2020} considered stochastic stability for \textsc{rl} in the unbounded state space setting. However, their work assumed a tabular setting, relied on access to the model of the environment, and ignored optimality to focus exclusively on stability. Our work makes stability practical for deep \textsc{rl} without having access to the environment transition dynamics and optimizes for both stability and optimality. The few other works that consider stability \citep{dai_ppo_2022, bai_rlqn_2022} assume that a stable policy is given and use it as a starting point for learning an optimal policy. We make no such assumption and try to learn a stable policy directly from a random policy. 

The combination of unbounded state space and the continuing setting is scarce in the \textsc{rl} literature. Existing works acknowledge the challenges from unboundedness, but they either artificially bound the state space \citep{bai_rlqn_2022, wei2023mixedsystem} or assume a finite-horizon training setting \citep{dai_ppo_2022}.

\section{Preliminaries}
\label{sec:prelim}

\textbf{Average-Reward}. Consider an infinite-horizon Markov decision process (\textsc{mdp}) \citep{puterman_mdp_2014}, $\mathcal{M} := \langle \mathcal{S}, \mathcal{A}, \mathcal{P}, c, d_1\rangle$, where $\mathcal{S}\subseteq \mathbb{R}^d$ is the state space, $\mathcal{A}$ the action space,  $\mathcal{P}:\mathcal{S}\times\mathcal{A}\to\Delta(\mathcal{S})$ the transition dynamics, $c:\mathcal{S}\times\mathcal{A}\times\mathcal{S}\to\mathbb{R}_{\geq 0}$ the cost function, and $d_1\in\Delta(\mathcal{S})$ the initial state distribution. Here we follow the control-theoretic convention and consider costs (the negation of rewards). Without loss of generality, we assume the cost is non-negative, and often restrict to cost-functions that are only dependent on the next state, $s_{t+1}$. Typically, and in the context of our work, the optimality cost function $c$ denotes a notion of distance from a target state such as the $L_1$ distance and the target state is the all-zero vector state (for more details, refer to Appendix~\ref{sec:more_env_details})~\citep{garcia_little_2008, ault_resco_2021}.

In the online task formulation, an agent, acting according to policy $\pi:\mathcal{S}\to\Delta(\mathcal{A})$, generates a \textit{single} infinitely long trajectory: $ s_1, c_1, a_1, s_2, ...$, where $s_1\sim d_1$, $a_t\sim\pi(\cdot|s_t), c_t = c(s_t,a_t,s_{t+1})$, and $s_{t+1}\sim \mathcal{P}(\cdot|s_t,a_t)$.  %
Unlike episodic RL, there are no resets in this formulation. Accordingly, we consider the long-run average-cost objective \citep{naik_discnotopt_2019, naik_continuing_2021}:
\begin{align}
    \label{eq:avg_cost_obj}
    \Jo(\pi) := \lim_{T\to\infty} \frac{1}{T} \sum_{t=1}^T  \E_\pi \left[ c_t \right].
\end{align}
An optimal policy is one that minimizes $\Jo(\pi)$. 
In the average-cost setting, the analog of the standard \textsc{rl} value functions are the \textit{differential value functions} which are defined as follows: the \textit{differential} action-value function, $Q^\pi(s, a) := \lim_{T\to\infty}\E_{\pi}[\sum_{t=1}^T (c_t - \Jo(\pi)) \mid s_1 = s, a_1 = a]$, the differential state-value function, $V^\pi(s) := \lim_{T\to\infty}\E_{\pi}[\sum_{t=1}^T (c_t - \Jo(\pi)) \mid s_1 = s]$, and the differential advantage function, $A^{\pi}(s,a) = Q^{\pi}(s,a) - V^{\pi}(s)$. The advantage function quantifies how good or bad an action is in a given state in yielding the long-term outcome. Note that since we are \emph{minimizing costs} rather than maximizing rewards, good and bad actions should have negative and positive advantage respectively.
We make the standard assumption that the \textsc{mdp} is communicating~\citep{bertsekas2015dynamic_vii}, meaning that for any pair of states $s$ and $s'$, there exists a policy that can transition from $s$ to $s'$ in a finite number of steps with non-zero probability. This assumption guarantees that the optimal average cost value $\inf_\pi \Jo(\pi)$ is independent of the starting state.

\textbf{Unbounded State Spaces and Stability}. In an unbounded state space, there is no limit to values that the features of a state can take on. This type of state space is different from the bounded state space formulation used in popular \textsc{rl} testbeds such as in MuJoCo  \citep{todorovmujoco2012} where all states are in a continuous \emph{bounded} region (e.g., robotic arm joint angles).

In unbounded state space stochastic control problems, \emph{stochastic stability} is a fundamental concept~\citep{shah_unbounded_2020}.
\begin{definition}[Stochastic Stability]
\label{def:stability}
A policy $\pi$ is stable if and only if the the average long-term incurred cost is bounded, i.e. $\Jo(\pi) < \infty$.
\end{definition}
Assuming nontrivial average cost $\inf_{\pi}\Jo(\pi) < \infty$ is possible, an optimal policy is stable. Since the cost $c$ can be a distance measure from some target state, a stable policy is one that achieves bounded distance from the target state.

\textbf{Lyapunov Functions}. Lyapunov functions are standard tools in control theory to analyze the stability of a system~\cite{meyn2012markov}. Intuitively, they measure the ``energy"  of a state where the energy of a state is typically directly proportional to the cost of being in that state, and the energy of the target state is zero. A \emph{control} Lyapunov function (\textsc{clf}), $\ell:\mathcal{S}\to\mathbb{R}_{\geq 0}$, is one that satisfies: 1) $\liminf_{\|s\|\to \infty} \ell(s) = \infty$ and 2) for all but a finite number of states $s$: $\min_{a\in\mathcal{A}}\E_{s'\sim \mathcal{P}(\cdot|s,a)}[\ell(s') - \ell(s)] < 0$ i.e. there exists an action that in expectation decreases the energy $\ell$ in \emph{one} step. It is known that finding a policy that outputs such actions will stabilize the system \cite{kelletstability2003}. However, it is difficult to determine such a function without knowledge of transition dynamics especially in highly stochastic settings. In these cases, it is common to resort to an approximate \textsc{clf}, where $\ell$ is decreased over multiple steps i.e. from a given state $s$, multiple actions are taken such that in the resulting state, $s''$, $\ell(s'') < \ell(s)$ \citep{taylorLearningLyp2019, NEURIPS2019_2647c1db, dai2021lyapunovstable}.

\section{Why is Achieving Stochastic Stability Difficult for Deep RL?}
\label{sec:deeprl_unbounded}
\begin{figure}[H]
  \centering
  {\includegraphics[scale=0.4]{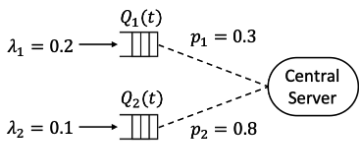}}
    \caption{\footnotesize $2$-queue network setting where a central server must select which of the two queues to serve. $Q_i$, $\lambda_i$, and $p_i$ are the queue length, arrival rate, and service probability of queue $i$ at each time-step. The image is taken from \citet{liu_rlqueueing_2019}.}
    \label{fig:2queue_example}
\end{figure}

In this section, we describe the core difficulties of achieving stochastic stability in highly stochastic \textsc{mdp}s with unbounded state spaces when minimizing the optimality cost objective. We do so using the queuing example in Figure~\ref{fig:2queue_example}.

 The queue lengths at time $t$ are given by $Q_i(t)$. The state is the vector of queue lengths i.e. $[Q_1(t), Q_2(t)]$, which is from an unbounded set since the queue lengths can grow arbitrarily large. The actions are discrete, denoting which queue the agent chose. In queuing theory, the optimality cost function is the total queue length of the system at the given time-step i.e. $c(s,a,s') = \|s'\|_1$ (see Appendix~\ref{sec:more_env_details} and \citet{garcia_little_2008}). An optimal policy is one that selects queues at each time-step such that the long-term average queue length is minimized. A stable policy ensures queue lengths are finite. In this specific example, the destabilizing action is one that serves an empty queue while the other queue is non-empty. If the agent too often chooses an empty queue, it runs the risk of the other queue blowing up in length. Thus, it is important that the agent quickly realizes that it should not serve the empty queue. Please refer to Appendix~\ref{sec:more_empirical} for an in-depth description of the environment. 

\textbf{Challenge 1: Poor Credit Assignment.} Credit assignment is one of the central challenges of \textsc{rl}. It is particularly, challenging in \emph{information-sparse} \textsc{mdp}s \citep{arumugam_creditinfo_2021, pignatelli2023CAPsurvey}, where due to high stochasticity in the transition dynamics, $\mathcal{P}$, a single action lacks influence on the \textsc{mdp}. Intuitively, this lack of influence arises when an action has unintended consequences due to the stochasticity of the system. For example, if the server successfully decreases the length of say $Q_1$, but then a new job simultaneously arrives then from the agent's perspective, it appears that the action failed. Note that this challenge is separate from the typical sparse $0/1$ cost signal widely cited in the \textsc{rl} literature \cite{pignatelli2023CAPsurvey,harutyunyan2019hindsight}; in our case, the cost function is dense, but the high stochasticity makes the \textsc{mdp} information-sparse.

In the queuing environment, a good action could have bad consequences (e.g., the server selects a non-empty queue, but due to stochasticity fails to serve it and a new job arrives) or a bad action could have neutral consequences (e.g., the server selects an empty queue and state remains the same). During learning, this noisiness can make distinguishing good actions from bad challenging since bad actions may appear good, and vice-versa. Ideally, serving an empty queue is adequately discouraged to avoid this confusion.

\textbf{Challenge 2: Poor Extreme Generalization.} The success of neural networks has primarily been due to their ability to \emph{locally} generalize when given a \emph{large} number of training samples per input~\citep{challot_deep_2017}. However, when the destabilizing actions are inadequately discouraged in the unbounded state space setting, the queue lengths may be blowing up, which results in the agent to drift away into new states $s$. This drift means that: 1) the neural network must perform \emph{extreme} generalization (e.g. generalize from queue lengths of $10$ to $10^4$) and 2) number of training samples for each previously-visited states $s$ is \emph{small}. In this situation, however, neural networks are known to output inaccurate predictions \citep{challot_deep_2017, xu_nnextra_2021, haley_nnextra_1992, barnard_nnextra_1992}. Thus, the inaccuracy of the agent's neural network inhibits the agent from performing effectively in its environment.

In Section~\ref{sec:empirical}, we show that online deep \textsc{rl} agents that do not handle these challenges perform poorly. In the next section, we present a cost shaping approach that explicitly encourages stability, which leads to discouragement of unstable actions, and a state transformation approach that makes deep \textsc{rl} algorithms more amenable to better generalization.

\section{STOP: STability and OPtimality}
\label{sec:learning_stab_opt}
We now introduce our method, STability and OPtimality (\textsc{stop}), an approach based on two ideas: 1) Lyapunov-based cost shaping that explicitly optimizes for stability in addition to optimality and 2) state transformations for better generalization across unbounded state samples. In Appendix~\ref{sec:pseudocode}, we include the pseudo-code.

\subsection{Lyapunov-based Cost Shaping}

Recall from Section~\ref{sec:prelim}, that an action that solves $\min_{a\in\mathcal{A}}\E_{s'\sim \mathcal{P}(\cdot|s,a)}[\ell(s') - \ell(s)] < 0$ leads to stabilizing the system. Our approach is to learn a policy that directly tries to do so. Thus, the agent minimizes the following stability and optimality objective:
\begin{align}
\label{eq:stability_objective}
\Js(\pi) & := \lim_{T\to\infty}\frac{1}{T}\sum_{t=1}^{T} \E_\pi [\underbrace{\ell(s_{t+1}) - \ell(s_{t})}_{\substack{\text{stability}}} + \underbrace{c(s_t, a_t, s_{t+1})}_{\text{optimality}}]
\end{align}

The Lyapunov function $\ell$ can take on many forms. In this section, we explain the intuition behind selecting $\ell$ using 
the 1) linear function: $\ell_1 := \|s\|_1$ and 2) quadratic form: $\ell_2 := \|s\|_2^2$ where $\|\cdot\|_{\{1,2\}}$ is the $\{1,2\}$-norm, and the true optimality cost is $c$. These choices are inspired by the literature on stochastic networks and control \citep{srikant_comm_2014, neely_networkcomm_2010}. As noted in Section~\ref{sec:prelim}, $\ell$ can be an approximate Lyapunov function---the above choices of $\ell$ are not exact in general. As shown in our experiments, an approximate Lyapunov function, could still offer significant benefits. In Section~\ref{sec:theory}, we draw a connection between $\Js$ and potential-based shaping~\citep{ngpotential1999}.

\subsubsection{Intuition for Encouraging Stability}
\label{sec:instability_intuition}
To understand why the Lyapunov-based shaping leads to stability, we analyze the advantage functions of destabilizing actions of a \textsc{ppo} agent in the queuing problem given in Figure~\ref{fig:2queue_example}. While in general it is unknown which actions destabilize the system, in the particular case of Figure~\ref{fig:2queue_example}, the destabilizing actions are those that serve empty queues when a non-empty queue is available. These actions should have positive and high advantage function estimates (recall we are minimizing costs). 

We analyze the normalized advantages of an average-reward \textsc{ppo} agent \citep{zhang_artrpo_2021, schulman_ppo_2017} at the \emph{start} of learning before any policy updates. Initially, the agent acts uniformly at random until it fills up the rollout buffer. The advantages are then computed using this data when the agent optimizes: 1) $c(s,a,s')$, 2) $\ell_1(s') - \ell_1(s) + c(s,a,s')$ (linear), and 3) $\ell_2(s') - \ell_2(s) + c(s,a,s')$ (quadratic). We can then compare the estimated normalized advantages for destabilizing actions. In Table~\ref{tab:adv_unstable}, we can see that with cost-shaping destabilizing actions are discouraged much more than without cost-shaping, which accordingly makes subsequent performance of the agent better. Without cost-shaping, destabilizing actions are assigned a relatively lower advantage contributing to less discouragement of destabilizing actions.

 \begin{table}[H]
 \vspace{-0.3cm}
\centering
\begin{tabular}{p{3.5cm}|p{4cm}}
\textbf{Cost Function} &
\textbf{Normalized Advantages} \\
$c(s,a,s')$ & $0.09 \pm 0.07$ \\ \hline
$c(s,a,s') + \ell_1(s') - \ell_1(s)$ & $0.25 \pm 0.08$ \\ \hline
$c(s,a,s') + \ell_2(s') - \ell_2(s)$ & $0.25 \pm 0.05$ \\ \hline
\end{tabular}
\vspace{-0.3cm}
\caption{Interquartile mean (\textsc{iqm}) statistics \citep{agarwal2021deep} of the normalized advantage estimates for unstable actions across $20$ trials in the $2$-queue example from Figure~\ref{fig:2queue_example}. Higher is better. To account for the variations in magnitudes, we divide the computed advantages by their standard deviation across the rollout buffer of size $128$. Note that since we are \emph{minimizing costs} rather than maximizing rewards, bad actions should have positive advantage.}
\vspace{-0.5cm}
\label{tab:adv_unstable}
\end{table}

\subsubsection{Choosing a Lyapunov Function}

In this section, we provide intuition on the choice of the fixed Lyapunov function by drawing from results on queuing theory and stochastic control. We first note that from Equation~\ref{eq:stability_objective}, we have it that the Lyapunov function is essentially used for potential-based reward shaping. \citet{ngpotential1999} note that the ideal potential function for reward shaping is the optimal value function i.e. $\ell(s) = V^*(s)$. However, the need for $V^*(s)$ creates a chicken-and-egg problem where to learn $V^*(s)$, we need an optimal policy, but we want to find the optimal policy. Thus, ideally we can resort to a Lyapunov function such that $\ell(s) \approx V^*(s)$. Moreover, intuitively, we would expect that since the stability cost $\ell(s') - \ell(s)$ models a hill-climbing strategy, we would expect \emph{any} \textsc{clf} $\ell$ to yield practical benefits. However, we empirically find that this expectation often fails.

Our selection for a Lyapunov function is inspired by the analysis of the popular MaxWeight (\textsc{mw}) algorithm in queuing \cite{stolyar_mw_2004}. There are variants of \textsc{mw} under a Lyapunov function of the form $\sum_i^n Q_i^{(\beta+1)}$ for $\beta>0$ \cite{stolyar_mw_2004}, where $\beta = 0$ corresponds to the linear Lyapunov function, and $\beta = 1$ corresponds to the quadratic function and classical \textsc{mw} algorithm. This algorithm has been proven to achieve a queue length that is at most within a factor of two from the optimal in certain heavy traffic regime of some queuing systems \cite{maguluri2016heavy}. For the \textsc{mw} variant with general $\beta$, the algorithm is stable but their delay performance is unclear compared to the classical choice of $\beta = 1$. Our selection is also based on the fact that using the linear Lyapunov function ($\beta = 0$) is difficult to ensure stability \cite{krishnasamy2019learning} and is typically used in very simple systems \cite{fayolle1993lyplinear}. Motivated by these prior analyses, we explore various values of $\beta$.

In Section~\ref{sec:empirical}, we explore a range of $\beta$ values, and find that as $\beta$ gets larger, the performance of the \textsc{rl} agent improves. However, once $\beta$ gets too large, performance deteriorates. Intuitively, this observation is expected: as we increase $\beta$, $\ell(s)$ starts to approximate $V^*(s)$ better based on the \textsc{mw} analysis; however, once $\beta$ becomes too large, the high variance of the large unbounded costs causes performance to degrade \cite{gupta2023behavior}. Therefore, when we use domain-knowledge to select $\ell$ in unbounded state space problems, we want $\ell$ such that it approximates $V^*(s)$ and its variance is not significantly high.

\subsubsection{Theoretical Properties}
\label{sec:theory}

We investigate the theoretical properties of Lyapunov-based cost shaping. All proofs are provided in Appendix \ref{sec:proof}.
The following proposition characterizes the relation between the reshaped objective $\Js$ and original optimality objective $\Jo$.

\begin{restatable}{proposition}{thmexpressionjs}
\label{thm:js_expression}
    For any policy $\pi$,
    \begin{align*}
        \Js(\pi) &= \Jo(\pi) + \lim_{T \to \infty} \frac{\E_\pi [\ell(s_{T+1})]}{T}.
    \end{align*}
    Therefore $\Js(\pi) \geq \Jo(\pi)$, and $\Js(\pi) = \Jo(\pi)$ if and only if $\E_\pi [\ell(s_{T+1})] = o(T)$.
\end{restatable}

The next proposition proves that under mild conditions, our approach is valid and will recover the optimal policy with respect to the original objective $\Jo$.
\begin{restatable}{proposition}{thmsameoptimalpolicy}
\label{thm:same_optimal_policy}
    Suppose that for any optimal policy $\pi^\star$, $\limsup_{T \to \infty} \E_{\pi^\star} [\ell(s_T)] < \infty$. Then $\argmin_\pi \Js(\pi) = \argmin_\pi \Jo(\pi)$.
\end{restatable}
The conditions for Proposition~\ref{thm:same_optimal_policy} are easily satisfied as long as the user-specified Lyapunov function $\ell$ is not growing too rapidly. For instance, if $\ell = \|\cdot\|_1$, then the above conditions hold when any optimal policy induces a stationary distribution and is stable (in the sense of Definition~\ref{def:stability}). If $\ell$ grows polynomially (e.g. $\ell = \|\cdot\|_2^2$), then it suffices for optimal policies to induce subexponential stationary distributions, which is true of a wide class of queuing problems and is ensured by standard technical assumptions \cite{hajek1982hitting, shah_unbounded_2020}. Proposition~\ref{thm:same_optimal_policy} is analogous to the main result of \citet{ngpotential1999}, which considers the discounted return criterion. They also make more stringent assumptions than ours, as in the unbounded state setting they require a uniformly bounded shaping term, which prohibits the use of Lyapunov functions.

\subsection{State Transformations}
\label{sec:state_transform}
\begin{figure}[H]
    \centering
    {\includegraphics[scale=0.4]{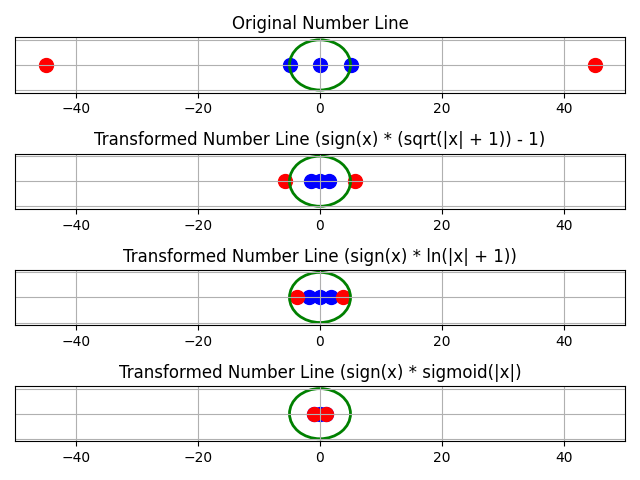}}
    \caption{\footnotesize State transformations applied to $1$D original untransformed points. Extreme points (red) appear closer to samples that agent was trained on (blue), thereby mitigating extreme generalization burdens. The green circle's radius is $5$ units and provides a sense of proximity of the points.}
    \label{fig:state_tr_intuition}
\end{figure}
In order to improve generalization in unbounded state spaces during training for more accurate credit assignment, we propose to use state transformations to compress the unbounded state space. These transformations are applied coordinate-wise to the state features, which are unbounded, before inputting them into the policy and value networks; the cost function is computed based on the original state.

Neural networks are known to perform \emph{extreme generalization} poorly \citep{challot_deep_2017, xu_nnextra_2021, haley_nnextra_1992, barnard_nnextra_1992}, which is problematic in the unbounded state space setting where states can be arbitrarily far from each other. Through Figure~\ref{fig:state_tr_intuition}, we provide some intuition on how state transformations can mitigate the burden of extreme generalization. Figure~\ref{fig:state_tr_intuition} shows how a set of points are transformed by three different transformation functions. The blue dots are points that the neural network has already trained on and the red extreme points are the points the neural network must generalize to.  We can see that in the original number line, the red points are ``far" from the blue points, which can make generalization challenging. However, when we apply different state transformations, we can shrink the distance between points while preserving ordering, thus mitigating the burden of extreme generalization.

We consider the following transformation functions: 1) symmetric square root: $\text{symsqrt}(x):= \text{sign}(x)(\sqrt{|x| + 1} - 1)$ \citep{kapturowski_sqrt_2018}, 2) symmetric natural log: $\text{symloge}(x):= \text{sign}(x)\ln(|x|+1)$ \citep{hafner_dreamerv3_2023}, and 3) $\text{symsigmoid}(x):=\text{sign}(x)(1/(1+e^{-|x|}))$. All these functions 1) are bi-jections and 2) preserve the ordering of points. Of course, there are trade-offs between transformations. For example, while symsigmoid reduces the generalization burden, it may effectively collapse the large states, making it challenging to differentiate between states and features. On the other hand, with symloge, the agent is still burdened by extrapolation but representation collapse is far less severe. 

\section{Empirical Study}
\label{sec:empirical}

We present an empirical study of \textsc{stop} on various challenging stochastic control tasks with unbounded state spaces.
We seek to answer the questions:
\begin{enumerate}[topsep=-1pt, itemsep=0pt, parsep=-1pt]
    \item Does \textsc{stop} enable learning of stable policies in settings where deep \textsc{rl} algorithms fail to?
    \item Are both the stability cost and state transformation components essential for learning stable policies?
\end{enumerate}

\subsection{Setup}

We first describe the environments, the algorithms we evaluate, and how we evaluate performance. We refer the reader to Appendix~\ref{sec:more_empirical} for more details.

\textbf{Environments.}
We conduct our experiments on the following environments. In each domain, the optimality cost function is the one typically used in the literature~\citep{garcia_little_2008, ault_resco_2021}.

\noindent
\textbf{Single-server allocation queuing:} We consider three variants of the $2$-queue setup: 1) medium load with no faulty connections (Figure~\ref{fig:2queue_example}), 2) high load with faulty connections, and 3) very high load with faulty connections. The optimality cost function is the average queue length and the target state is all-zero queue lengths.

\noindent
\textbf{$N$-model network:} We consider three setups of the $N$-network model \cite{harrisonNmodel1998} : 1) high load, 2) very high load $\#1$, and 3) very high load $\#2$. The optimality cost function is the average holding cost and the target state has $0$ holding cost.

\noindent
\textbf{Traffic control:} We also evaluate our approach on the \textsc{sumo} \citep{behrisch_sumo_2011} traffic simulator, but due to space constraints defer to Appendix~\ref{sec:more_empirical}. In this domain, \textsc{sumo} models a real-life traffic situation by bounding the number of cars per lane. However, the success of \textsc{stop} and failure of existing deep \textsc{rl} algorithms suggests our ideas are even applicable to the \emph{bounded} state space setting. The optimality cost function is the total waiting time and the target state has $0$ waiting time.

\textbf{Algorithms.}
Our baseline is average-reward \textsc{ppo} \citep{zhang_artrpo_2021} (denoted by \textsc{o} since it optimizes the optimality criterion only) since it is designed: 1) for the infinite horizon setting without discounting \citep{naik_continuing_2021} and 2) to be robust to hyperparameter tuning.  For the server allocation and $N$-model networks environments, we also evaluate  \textsc{maxweight} \citep{tassiulas_mw_1990}, a strong baseline algorithm with knowledge of the transition dynamics. In particular, \textsc{maxweight} is known to be asymptotic optimal in heavy-traffic regimes \cite{maguluri2016heavy}. When \textsc{ppo} is equipped with our cost-shaping approach and/or state transformations, we refer to it as \textsc{stop}.  \textsc{mw} starts the online interaction process with perfect estimation of some part of the transition dynamics, while \textsc{stop} must learn a stable and optimal policy without this knowledge. Note that: 1) the optimal policy is unknown in our environments~\cite{ganti_qopen-prob_2007, dai_ppo_2022}, 2) it is generally unknown how far \textsc{maxweight} is from optimal \citep{tassiulas_mw_1990}, and 3) computing the optimal policy with full knowledge of the transition dynamics is not straightforward since there are infinite states in the unbounded state setting~\citep{shah_unbounded_2020}.

\textbf{Online Evaluation.}
The agent starts from a random start state with a randomly-initialized policy and is never reset. We plot the true cost incurred by the agent vs. interaction time-steps. An increasing curve indicates instability as $c(s,a,s')\to\infty$, so the agent is drifting away from the target state; a flat curve indicates stability as $c(s,a,s')<\infty$; and a decreasing curve indicates improvement towards optimality as $c(s,a,s')\to0$.

\begin{figure*}[hbtp]
    \centering
        \subfigure[Fraction of unstable actions taken] {\includegraphics[scale=0.12]{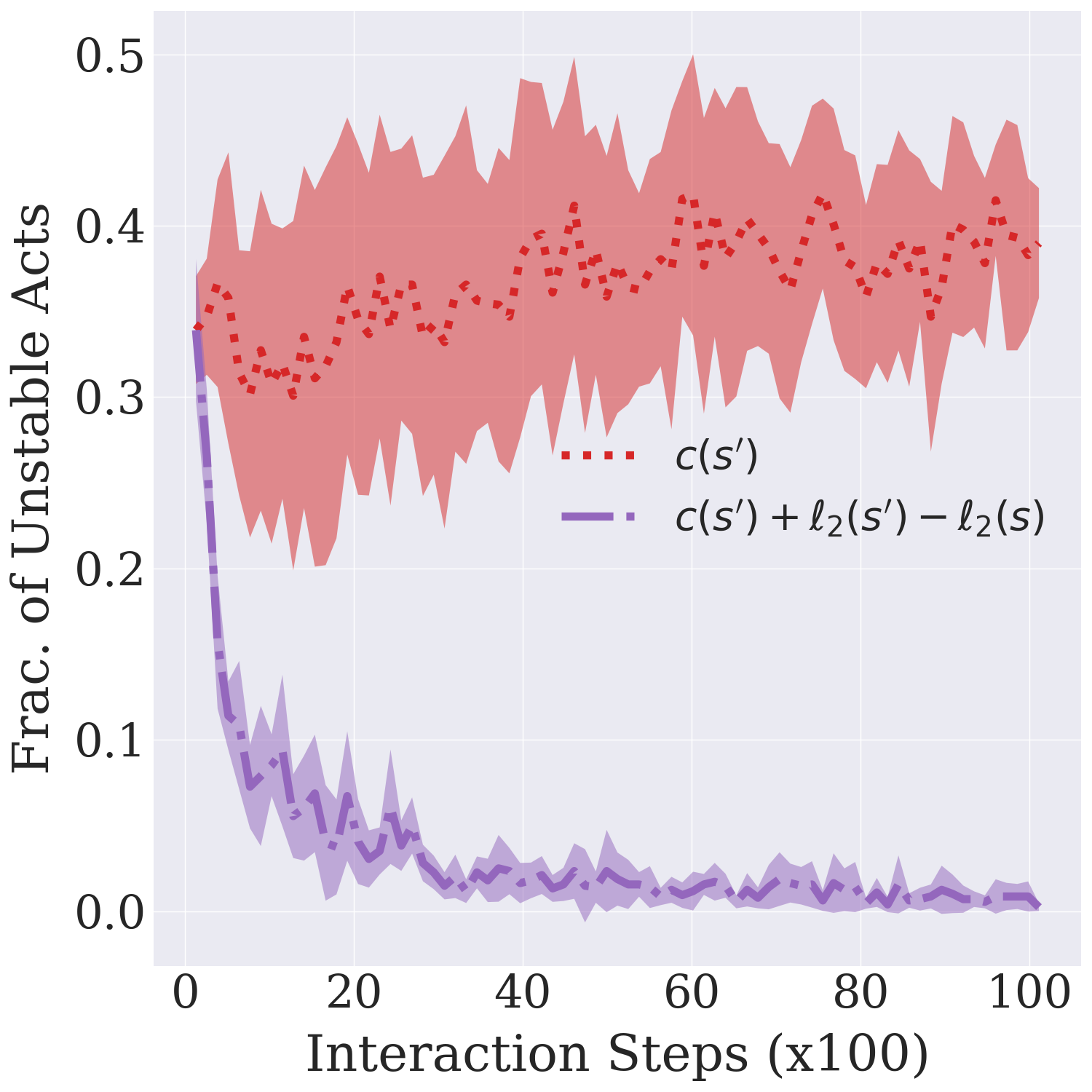}}
        \subfigure[\textsc{stop} state visitation distribution]{\includegraphics[scale=0.37]{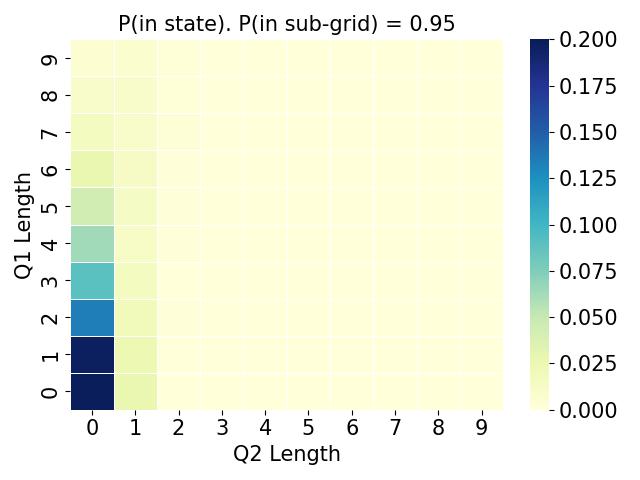}}
        \subfigure[\textsc{ppo} state visitation distribution]{\includegraphics[scale=0.37]{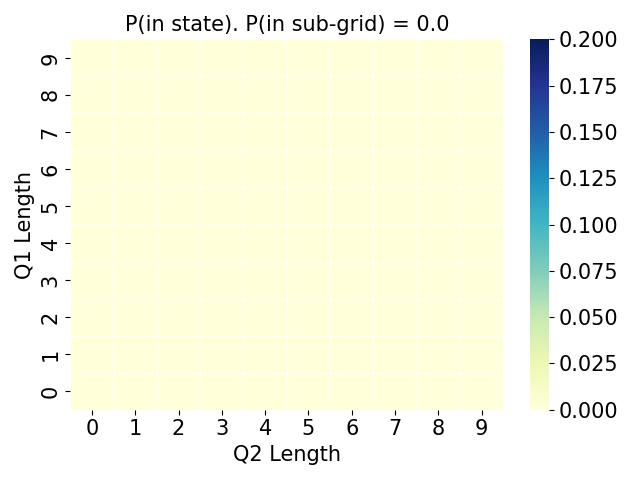}}
    \caption{\footnotesize Stability verification. (a) fraction of unstable actions taken by the agent over the course of training of \textsc{stop} ($c(s') + \ell_2(s') - \ell_2(s)$) and \textsc{ppo} $(c(s'))$ agents; lower is better. (b) and (c) state-visitation distribution of \textsc{stop} and \textsc{ppo} agents respectively; higher $\Pr(\text{in sub-grid})$ is better. Note that the empty region of (c) shows the failure of the \textsc{ppo} agent to visit the specified bounded region near the target state. All quantities were computed over $10$ trials on the $2$-queue setting from Figure~\ref{fig:2queue_example}.}
    \label{fig:stab_ablation}
\end{figure*}
\begin{figure*}[hbtp]
    \centering
        \subfigure[No faulty connections with medium load (Figure~\ref{fig:2queue_example})]{\includegraphics[scale=0.13]{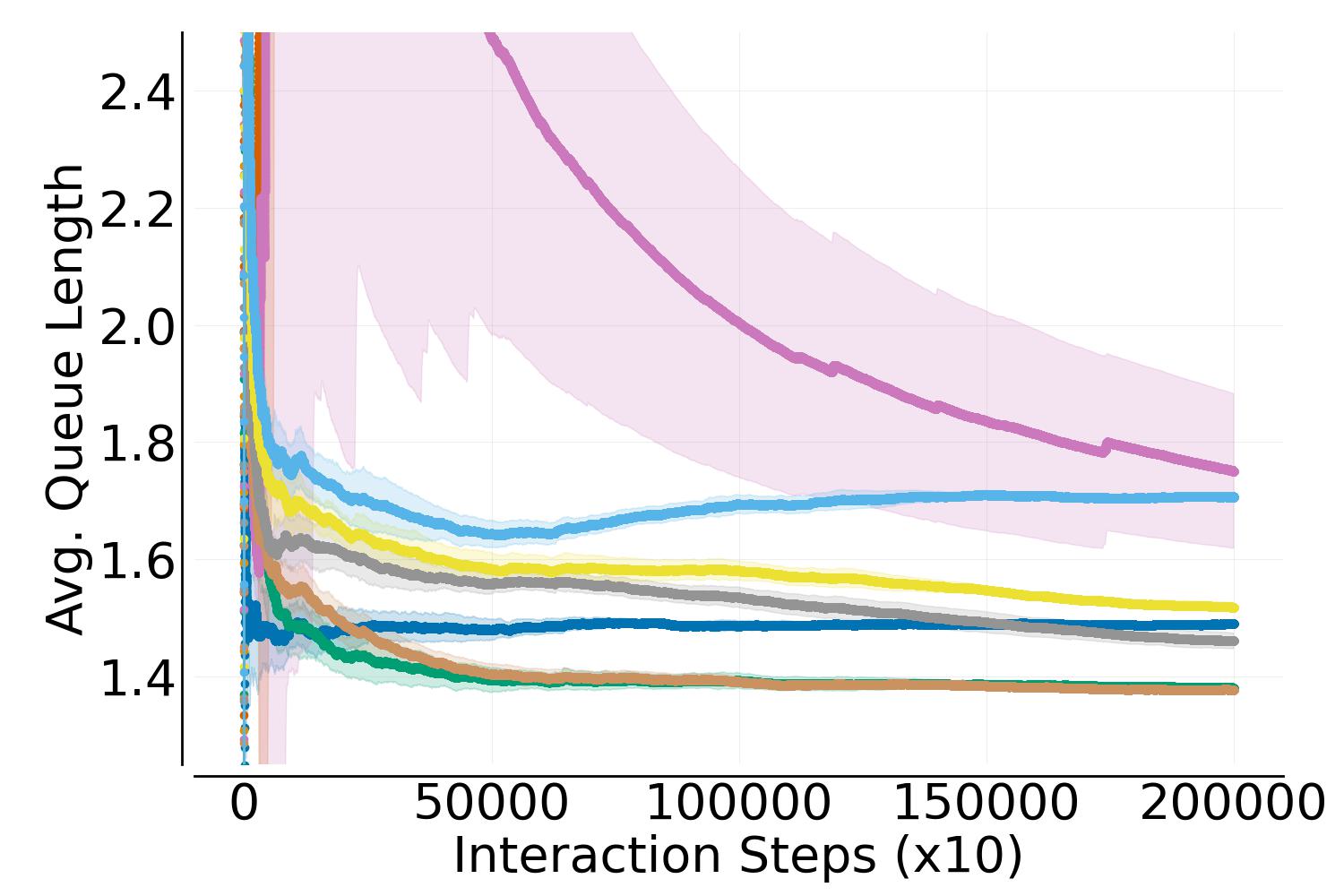}}
        \subfigure[Faulty connections with high load]{\includegraphics[scale=0.13]{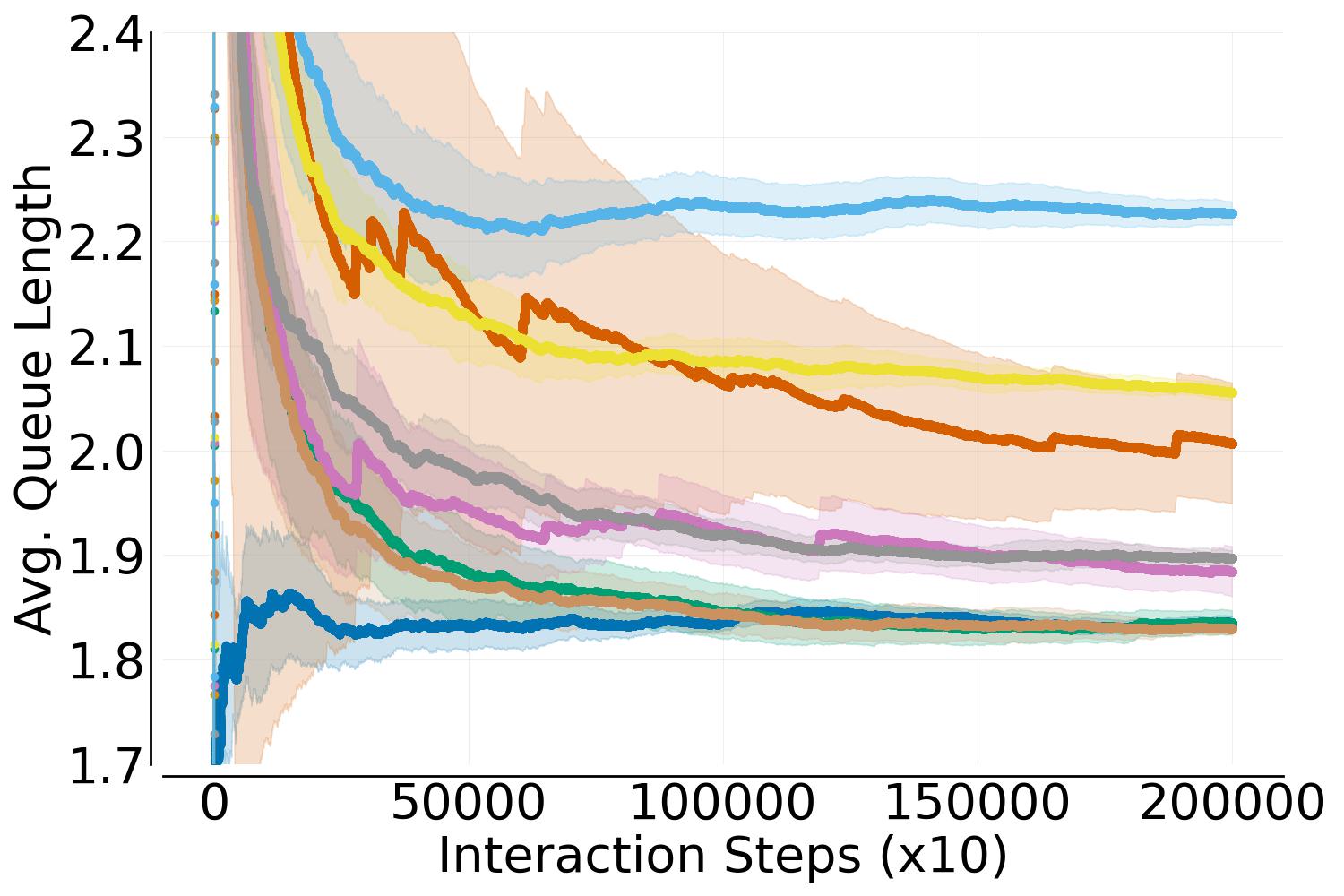}}
        \subfigure[Faulty connections with very high load]{\includegraphics[scale=0.13]{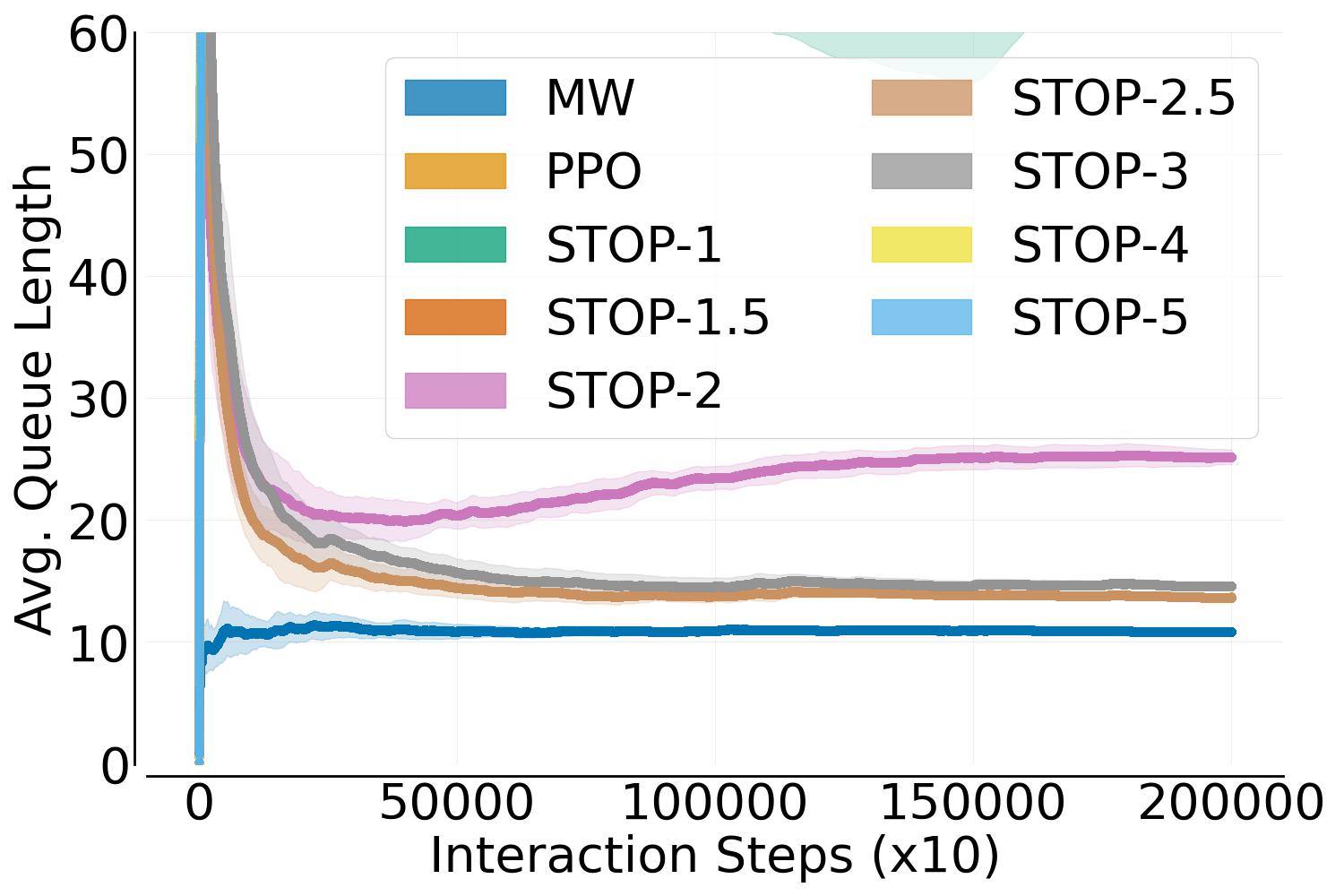}}\\
        \subfigure[High Load]{\includegraphics[scale=0.13]{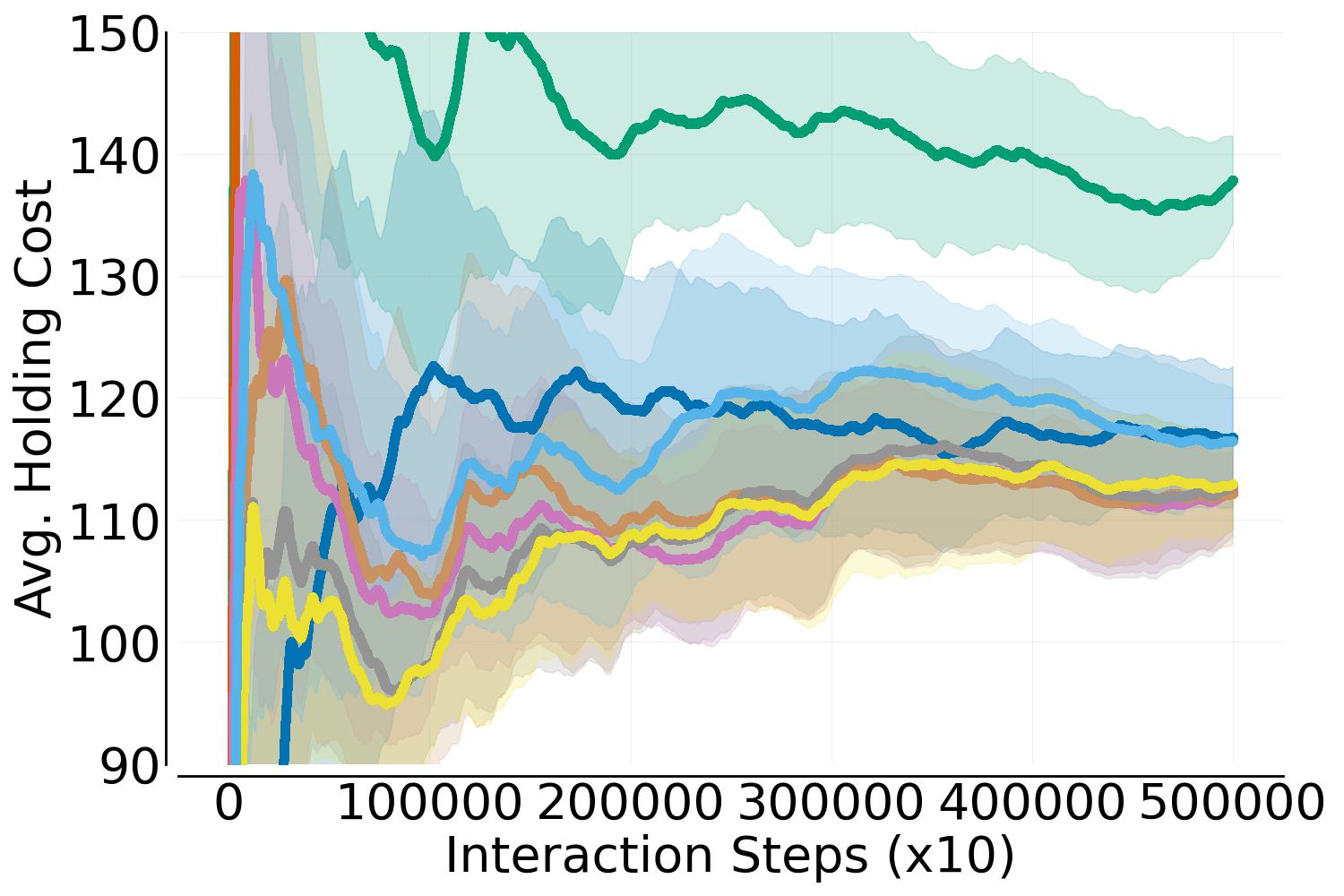}}
        \subfigure[Very High Load - $\#1$]{\includegraphics[scale=0.13]{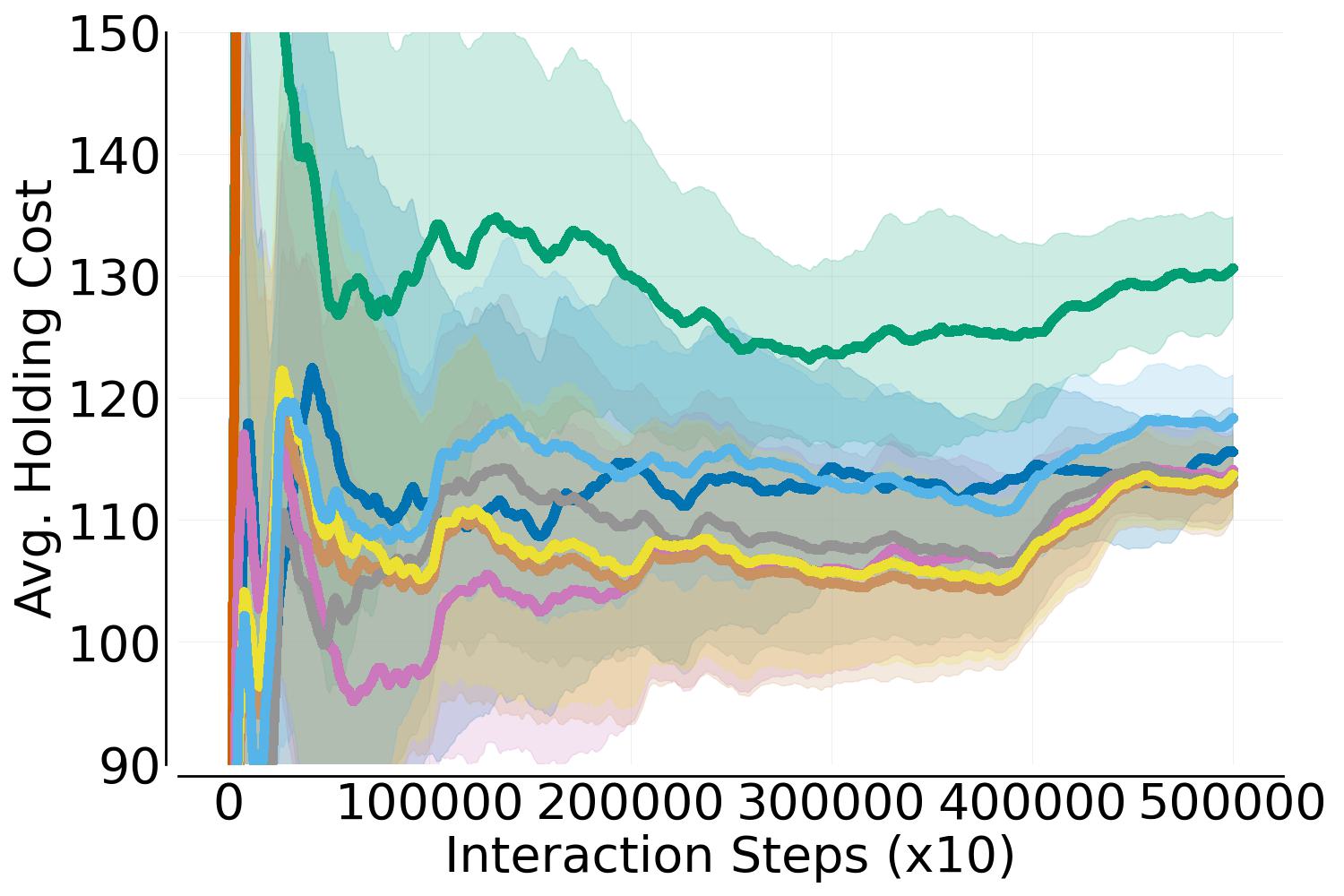}}
        \subfigure[Very High Load - $\#2$]{\includegraphics[scale=0.13]{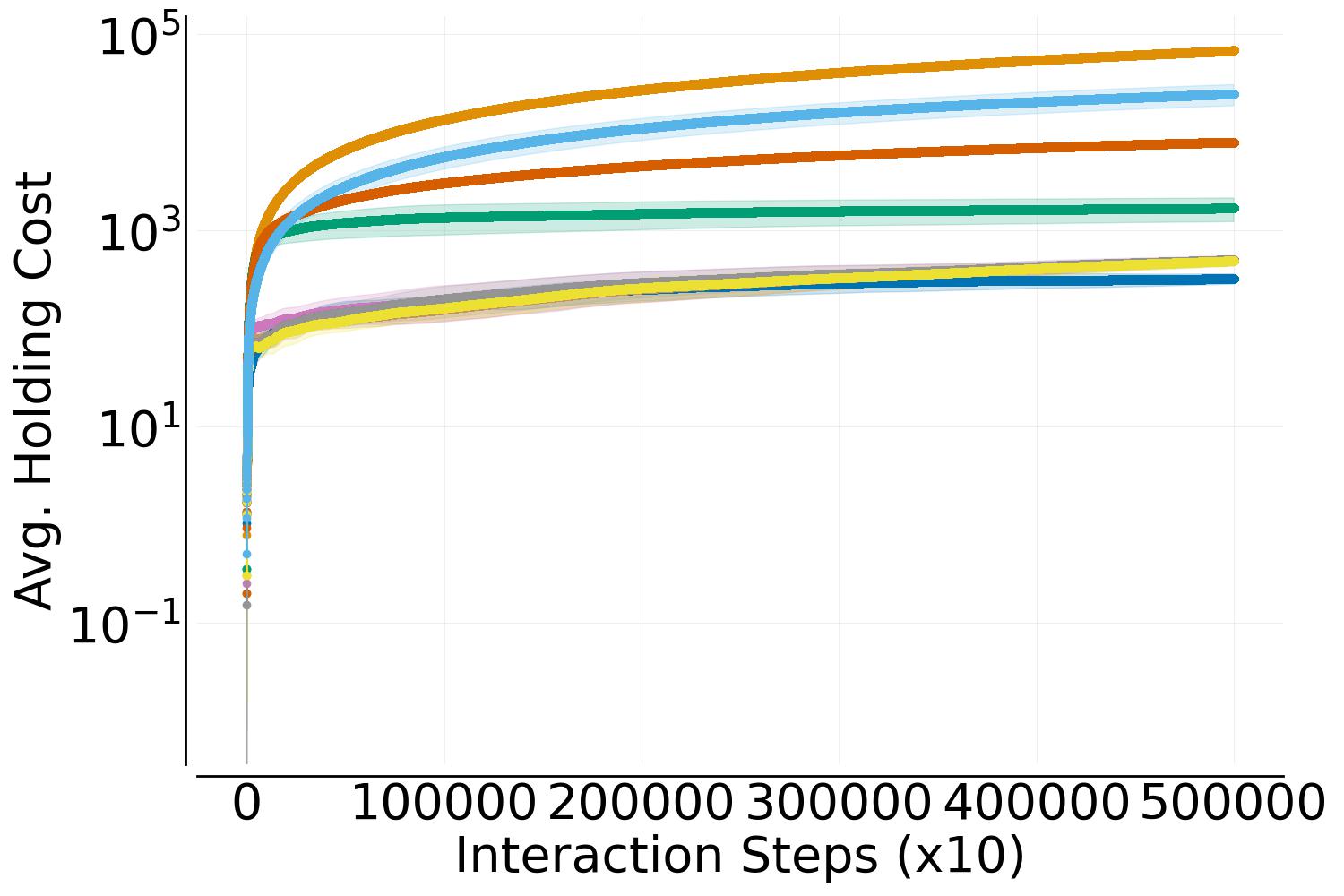}}
    \caption{\footnotesize True optimality criterion vs. interaction time-steps on three server-allocation queue networks (top row) and three $N$-network model environments (bottom row). Lower is better. Algorithms are \textsc{ppo} vs. \textsc{stop}-$p$, where $p$ denotes the power of the Lyapunov function. All \textsc{stop} variants use the symloge state transformation.  We also report the performance of \textsc{maxweight} (\textsc{mw}). Recall that unlike \textsc{mw}, \textsc{stop} does not know the transition dynamics. The \textsc{iqm} \cite{agarwal2021deep} is computed of the performance metrics over $20$ trials with $95\%$ confidence intervals. The vertical axis of (f) is log-scaled. To enhance visibility we zoom into the plot, which hides performance of \textsc{o} in some cases since it was unstable. We refer the reader to Appendix~\ref{sec:more_empirical} for the zoomed-out plots.} 
    \label{fig:base}
\end{figure*}
\subsection{Stability Encouragement}
Our first set of experiments aims to determine whether \textsc{stop} stabilizes the agent. We answer two questions: 1) does the number of destabilizing actions taken reduce over training? and 2) is the \textsc{stop} agent within a low bounded $L_1$ distance from the all-zero queue lengths with high probability?

In answering both these questions, we use the quadratic Lyapunov function, $\ell_2(s) =\|s\|_2^2$ for our shaping cost with no state transformations on the queuing network shown in Figure~\ref{fig:2queue_example}. We answer the first question with Figure~\ref{fig:stab_ablation}(a) where we plot the fraction of destabilizing actions taken by the \textsc{ppo} agent when it is optimizing with (\textsc{stop}) and without the shaped cost. We calculate destabilizing actions as the fraction of $(s,a)$ samples in the rollout buffer which serve an empty queue. We can see that the \textsc{stop} quickly stops taking destabilizing actions, while $\approx 40\%$ of the actions taken by \textsc{ppo} without cost-shaping are destabilizing.

We answer the second question with Figure~\ref{fig:stab_ablation}(b) and (c). In Figure~\ref{fig:stab_ablation}(b) and Figure~\ref{fig:stab_ablation}(c), we plot the state-visitation distribution of the agent after $100$K interaction steps. We see that the \textsc{stop} agent visits states that have an $L_1$ distance of at most $20$ with probability $0.95$, while the vanilla \textsc{ppo} agent does not even appear in this bounded region.

\subsection{Main Results}

We now compare vanilla \textsc{ppo} algorithm to \textsc{ppo} equipped with \textsc{stop} in Figure~\ref{fig:base} on a variety of highly stochastic \textsc{mdp}s. We set the rollout buffer length to $200$ and keep all other hyperparameters for \textsc{stop} and the baseline the same \citep{huang2022cleanrl}, and show results for agents that achieved the lowest true optimality cost at the end of interaction time. We evaluate the following stability variations: $\ell(s)=\|s\|_p^p$, where $p = \{1, 1.5, 2, 2.5, 3, 4, 5\}$. All \textsc{stop} agents use the $\text{symloge}$ transform.

In all experiments, the na\"ive \textsc{ppo} agents, which optimized the optimality criterion directly and used no state transformations, were unstable. On the other hand, \textsc{stop} agents are able to achieve stability. In $4/6$ cases, we found \textsc{stop} out-performed or was extremely competitive with \textsc{mw}. To the best of our knowledge, no online \textsc{rl} algorithm in unbounded state spaces has outperformed \textsc{mw}. In very high load environments, such as Figure~\ref{fig:base}(f) we find that the high load causes all algorithms to be unstable.

In the highly stochastic queuing environments only, we found that using the linear stability cost and optimality cost is insufficient to achieve stability: the linear stability cost (bounded between $[-1,1]$) and optimality cost have different magnitude scales, which can cause the latter to dilute the former, effectively eliminating any benefit of the stability cost. In these cases, we replaced the optimality cost $c$ with $-1/(c + 1)$. On the other hand, we were able to keep the original optimality cost for all the other variations. While the linear agent ($p = 1$) can stabilize the system, the use of the reciprocal may contribute to the relatively poor performance compared to the agents ($p \neq 1$).

In general, we found that moderate values of $p$ achieved the lowest average queue length or holding cost in high load settings. For example, in $p=3$ (cubic; gray line) performed reasonably well in all the environments. While $p=1$ (linear; green line) and $p=5$ (light blue line) performed well in only two or three environments (see: (a), (b), (d), (e)). The ability of $p=1$ to achieve low average queue length in settings (a) and (b) corresponds to the fact that the policy that serves the queue with the largest service rate ($c\mu$-rule) performs well in these settings \cite{Buyukkoc1985TheCR}. These results align with our expectation: smaller values of $p$ may inadequately discourage de-stabilizing actions and are potentially poorer approximations of the optimal value function, while larger $p$ values result in high variance of the unbounded cost, which can degrade performance.

\subsection{Ablation Studies}
We have shown that \textsc{stop} enables learning in highly stochastic environments with unbounded state spaces. We now analyze the importance of different components of \textsc{stop}. 

\begin{figure}[H]
    \centering
        \subfigure[\textsc{stop} with varying state transformations]{\includegraphics[scale=0.105]{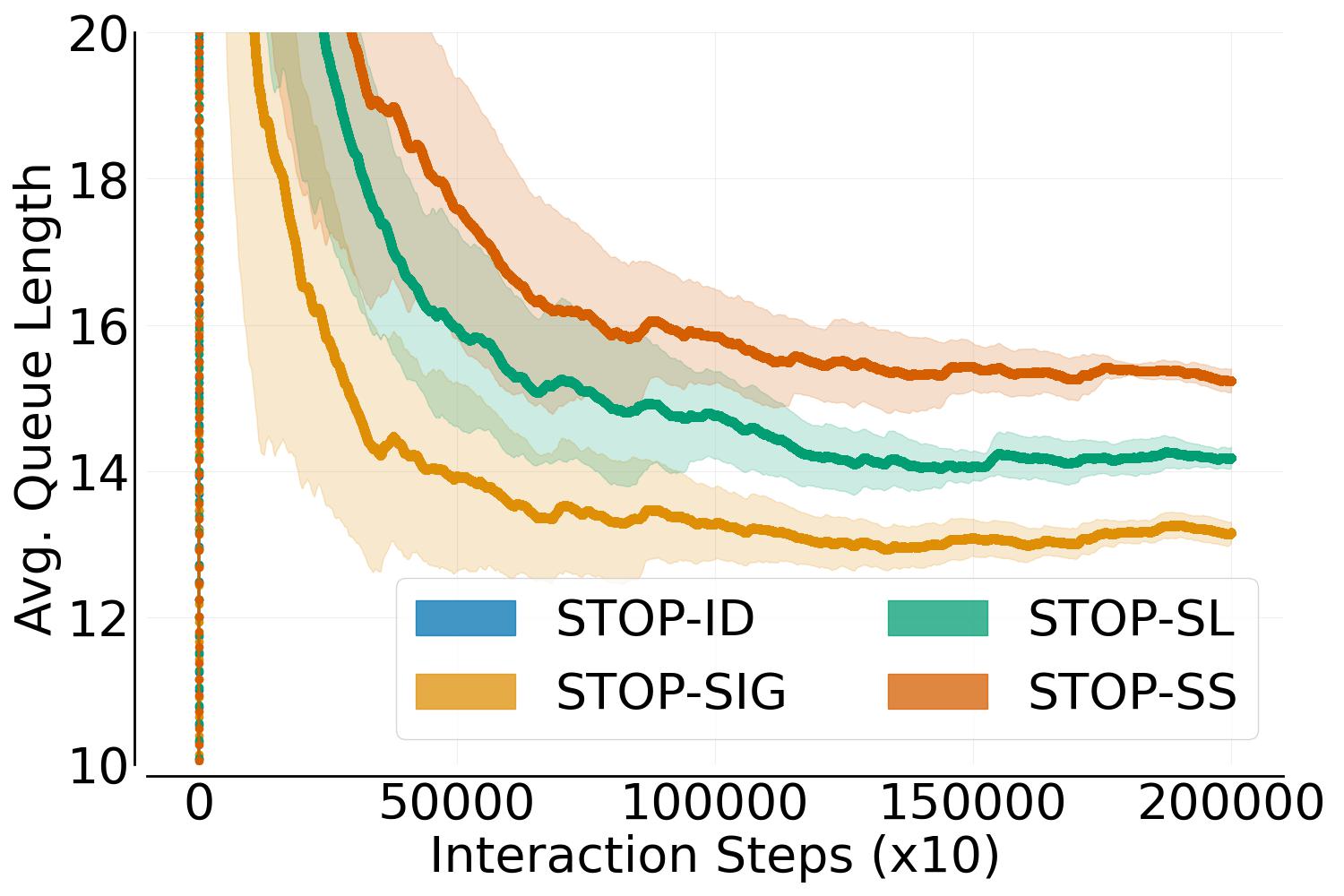}}
        \subfigure[Only state transformations, no stability cost]{\includegraphics[scale=0.105]{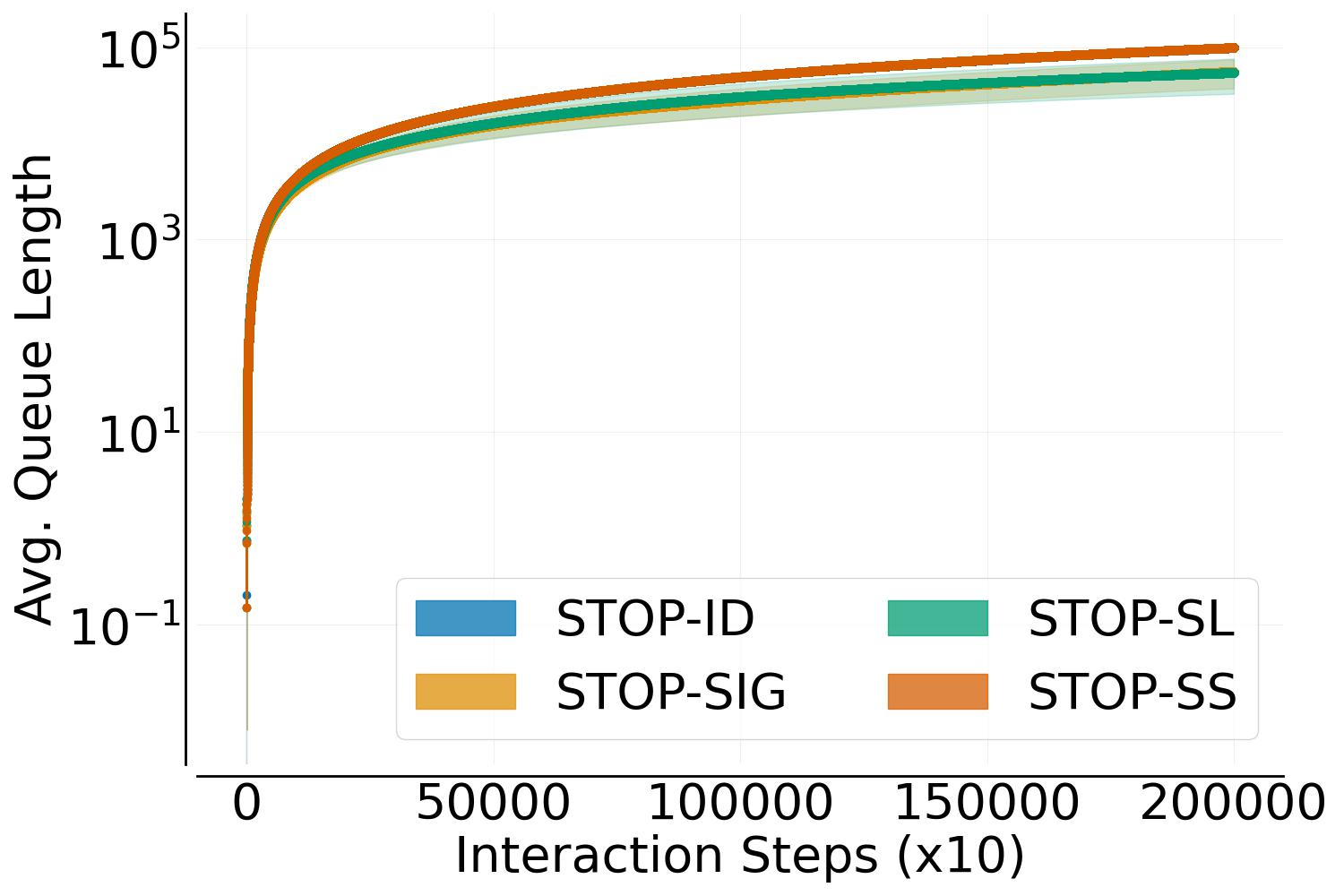}}
        
    \caption{\footnotesize Ablations. True optimality criterion vs. interaction time-steps on $2$ queue setup with faulty connections. Lower is better. The \textsc{iqm} \cite{agarwal2021deep} is computed of the performance metrics over $20$ trials with $95\%$ confidence intervals. The vertical axis of (b) is log-scaled. \textsc{id}: identity; \textsc{sig}: symsigmoid, \textsc{ss}: symsqrt; \textsc{sl}: symloge. Refer to Appendix~\ref{sec:more_empirical} for the zoomed-out plots.}
    \label{fig:ablation}
\end{figure}

\subsubsection{STOP with Varying State Transformations}

In this experiment, we seek to understand the importance of the state transformation component of \textsc{stop}. We evaluate the performance of a \textsc{stop} \textsc{ppo} agent with $\ell(s)=\|s\|_3^3$  (cubic) and different state transformations from Section \ref{sec:state_transform}. From Figure~\ref{fig:ablation}(a), applying state transformations result in significantly better performance compared to no application of such a transformation. The significant improvement over no state transformations (\textsc{id}) suggests that state transformations are indeed critical and that in very difficult environments the stability cost is insufficient. For different transformations, the agent faces different levels of extreme generalization burden. When the agent uses sigmoid, all its extreme generalization burden is mitigated, while when it uses the squareroot it suffers more of the burden compared to log and sigmoid.

\subsubsection{State Transformations with No Stability Cost}
In Figure~\ref{fig:ablation}(b), we remove the stability cost. We observe that applying only state transformations is insufficient for the agent to stablize and the stability cost component of \textsc{stop} is critical. This result also confirms that simply bounding the state space (such as applying sigmoid) is insufficient to stabilize the system.

\section{Conclusion}
Our work showed that online deep \textsc{rl} agents that directly try to be optimal in highly stochastic environments with unbounded state spaces actually end up unstable. To address this instability, we introduced \textsc{stop} that explicitly encourages stability and optimality. We provided insight on how to leverage domain knowledge from queuing theory to appropriately select Lyapunov functions that: 1) approximate the optimal value function, 2) adequately discourage de-stabilizing actions, and 3) have low variance. We showed that \textsc{stop} trained highly performant deep \textsc{rl} policies in these difficult environments, and even out-performed strong baselines from the queuing literature.  \textsc{stop} improves the reliability of online deep \textsc{rl} policies on challenging stochastic control problems by ensuring that the policies are within bounded distance from the desired state.

A key takeaway is that deep \textsc{rl} algorithms that are successful on MuJoCo and Atari may fail to generalize well to environments with high stochasticity and unbounded state spaces. We shed light on the fact that there are real-world-inspired problems that appear simple, but are surprisingly challenging, and that as a community we must innovate new techniques to solve these challenging problems.

In this work, we showed that dense optimality cost functions used in real-world-inspired domains could be poor learning signals. An interesting future direction would be to formally investigate why they are so beyond their inability to do proper credit assignment. Another interesting future direction would be to explore how we can learn an appropriate Lyapunov function from data.

\section*{Impact Statement}
Our work is largely focused on studying fundamental \textsc{rl} research questions, and thus we do not see any immediate negative societal impacts. The aim of our work is to improve the reliability of deep \textsc{rl}  algorithms on real-world-inspired domains. By leveraging \textsc{stop}, a user takes a step towards improving the reliability of their decision-making agents.

\section*{Acknowledgements}
The authors thank Abhinav Narayan Harish, Adam Labiosa, Andrew Wang, Lucas Poon, and the anonymous reviewers at NeurIPS, ICLR, and ICML for their feedback on earlier drafts of this work. Y.\ Chen and M.\ Zurek were supported in part by National Science Foundation Awards CCF-2233152 and DMS-2023239. Q.\ Xie was supported in part by National Science Foundation Awards CNS-1955997 and EPCN-2339794.
J.\ Hanna and B.\ Pavse were supported in part by American Family Insurance through a research partnership
with the University of Wisconsin—Madison’s Data Science Institute.

\nocite{langley00}

\bibliography{main}
\bibliographystyle{icml2024}

\newpage
\appendix
\onecolumn
\section{Theoretical Results}
\label{sec:proof}

In this section we provide the proofs for our theoretical results, which are restated below for readers' convenience.

\thmexpressionjs*
\begin{proof}
Fix a policy $\pi$. Then we can calculate that
\begin{align*}
    \Js(\pi) &= \lim_{T \to \infty} \frac{1}{T} \E_\pi \left[\sum_{t=1}^T \left(c(s_t, a_t, s_{t+1}) + \ell(s_{t+1}) -\ell(s_{t}) \right)\right] \\
    &= \lim_{T \to \infty} \frac{1}{T} \E_\pi \left[\ell(s_{T+1}) -\ell(s_{1}) + \sum_{t=1}^T c(s_t, a_t, s_{t+1}) \right] \\
    &= \lim_{T \to \infty} \frac{1}{T} \E_\pi \left[\ell(s_{T+1})\right]  + \frac{1}{T} \E_\pi \left[\sum_{t=1}^T c(s_t, a_t, s_{t+1}) \right] \\
    &= \Jo(\pi) + \lim_{T \to \infty} \frac{1}{T} \E_\pi \left[\ell(s_{T+1})\right]
\end{align*}
where we used the fact that $\lim_{T \to \infty} \frac{\ell(s_1)}{T} = 0$.

The fact that $\Js(\pi) \geq \Jo(\pi)$ then follows from the fact that $\ell$ is non-negative. 
\end{proof}

\thmsameoptimalpolicy*
\begin{proof}
Fix a policy $\pi^\star$ which is optimal for $\Jo$. From Proposition~\ref{thm:js_expression} we have that $\Js(\pi^\star) \geq \Jo(\pi^\star)$. To show the reverse inequality, we can use Proposition~\ref{thm:js_expression} to calculate that
\begin{align*}
    \Js(\pi^\star) &= \Jo(\pi^\star) + \lim_{T \to \infty} \frac{1}{T} \E_{\pi^\star} \left[\ell(s_{T+1})\right] \\
    & \leq \Jo(\pi^\star) + \limsup_{T \to \infty} \frac{1}{T}\E_{\pi^\star}  [\ell(s_T)] \\
    & = \Jo(\pi^\star)
\end{align*}
using the assumption that $\limsup_{T \to \infty} \E_{\pi^\star} [\ell(s_T)] < \infty$. Therefore $\Js(\pi^\star) = \Jo(\pi^\star)$. Now for any other policy $\pi$, since $\pi^\star$ is optimal for $\Jo$, we have that $\Jo(\pi^\star) \leq \Jo(\pi)$. Therefore
\[\Js(\pi^\star) = \Jo(\pi^\star) \leq \Jo(\pi) \leq \Js(\pi)\]
where we used the fact from Proposition~\ref{thm:js_expression} that $\Js(\pi) \geq \Jo(\pi)$. Therefore $\pi^\star \in \argmin_\pi \Js(\pi)$, so $\argmin_\pi \Js(\pi) \supseteq \argmin_\pi \Jo(\pi)$. Furthermore, if $\pi \not \in \argmin_\pi \Jo(\pi)$, then we must have $\Jo(\pi) > \Jo(\pi^\star)$, in which case by again using Proposition~\ref{thm:js_expression} we have that
\[\Js(\pi) \geq \Jo(\pi) > \Jo(\pi^\star) = \Js(\pi^\star)\]
so $\pi \not \in \argmin_\pi \Js(\pi)$. Thus $\argmin_\pi \Js(\pi) \subseteq \argmin_\pi \Jo(\pi)$ and we can conclude that $\argmin_\pi \Js(\pi) = \argmin_\pi \Jo(\pi)$ as desired.
\end{proof}

\section{STOP Pseudo-code}
\label{sec:pseudocode}

 \begin{algorithm}[H]
  \caption{\textsc{stop+ppo}}
  \begin{algorithmic}[1]
    \STATE Input: policy parameters $\theta_0$, critic net parameters $\phi_0$, state transformation function $\sigma$, rollout buffer $\mathcal{D}$ of length $N$.
    \FOR{t = 1, 2, ...}
        \STATE Collect sub-trajectory in rollout buffer $\{\sigma(s_k), a_k, \sigma(s_{k+1}), l_k\}_{k=1}^{N}$ from environment using $\pi_{\theta_{\lfloor t / N\rfloor}}$ \COMMENT{Note that rollout buffer contains the transformed states and the cost $l_k := \underbrace{\ell(s_{k+1}) - \ell(s_{k})}_{\substack{\text{stability cost}}} + \underbrace{c(s_k, a_k, s_{k+1})}_{\text{optimality cost}}$ is a function of the non-transformed states.}
         \IF {$t \% N == 0$} 
         \STATE \COMMENT{Periodically update policy and critic parameters}
            \STATE Using rollout buffer $\mathcal{D}$ update $\theta$ and $\phi$ with average-reward PPO \cite{zhang_artrpo_2021}.
            \STATE Empty $\mathcal{D}$
         \ENDIF
         \STATE Record performance of agent according to true optimality cost at time-step $t$, $c(s_t, a_t, s_{t+1})$, as a function of non-transformed states $\{s_t,s_{t+1}\}$.
    \ENDFOR
  \end{algorithmic}
\end{algorithm}

\section{Supporting Content and Empirical Results}
\label{sec:more_empirical}

In this section, we include additional details and experiments that complement the main results. We also include the code in the supplementary zip file.

\subsection{Visualizations of State Transformations}

\begin{figure*}[h]
  \begin{center}
    \includegraphics[width=0.3\textwidth]{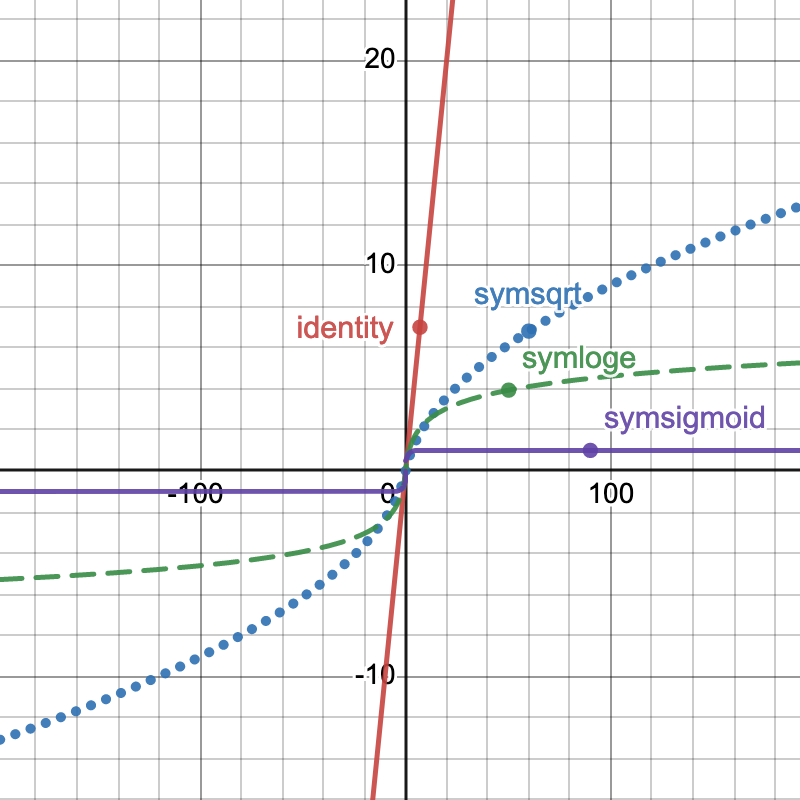}
  \end{center}
  \caption{\footnotesize Visualizations of transformation functions.}
  \label{fig:state_transforms}
\end{figure*}

To provide better intuition of the different state transformations we considered in Section \ref{sec:state_transform}, we visualize them in Figure \ref{fig:state_transforms}.

\subsection{Additional Main Results}

\begin{figure*}[h]
    \centering
        \subfigure[Faulty connections with medium load (Figure~\ref{fig:2queue_example})]{\includegraphics[scale=0.13]{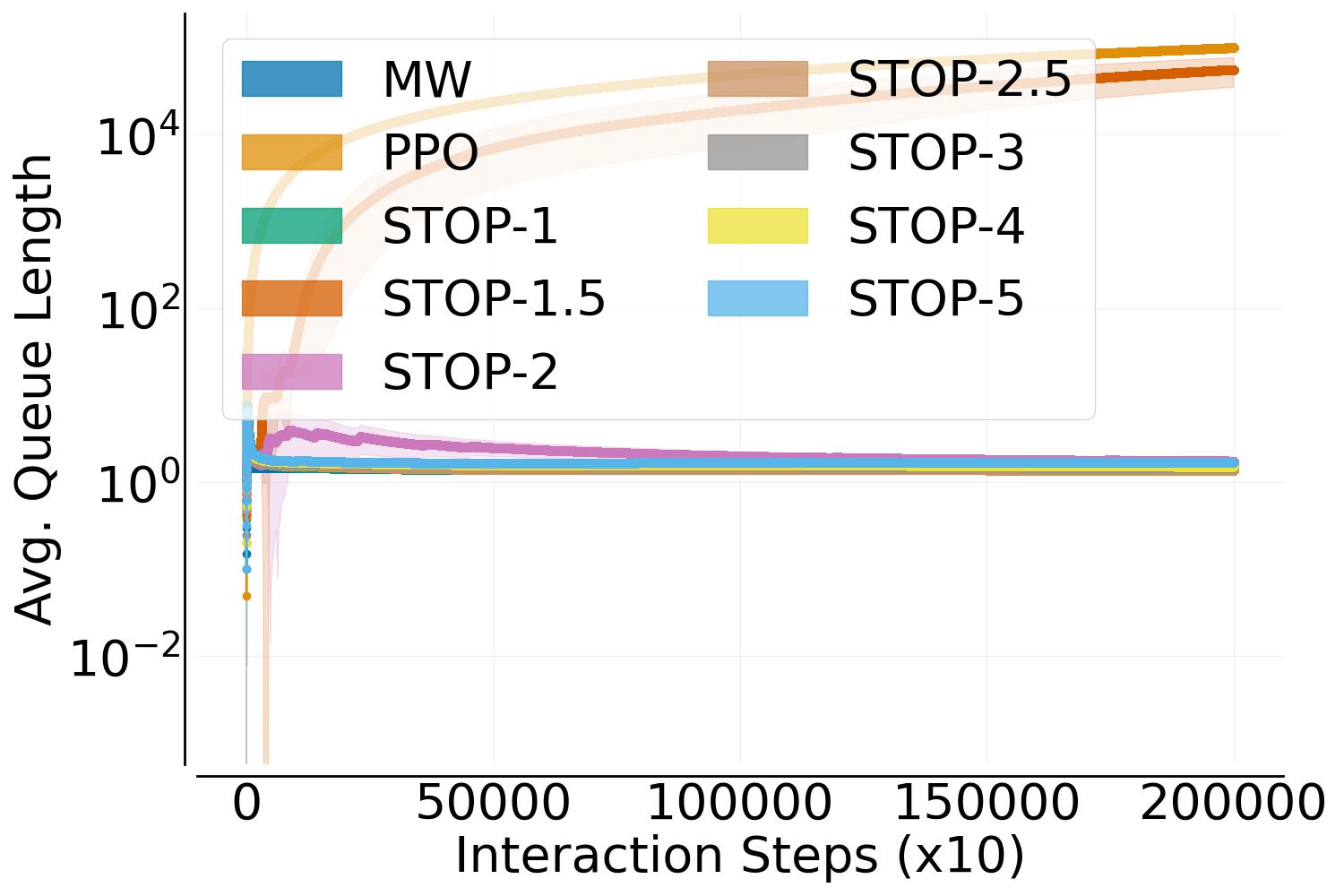}}
        \subfigure[Faulty connections with high load]{\includegraphics[scale=0.13]{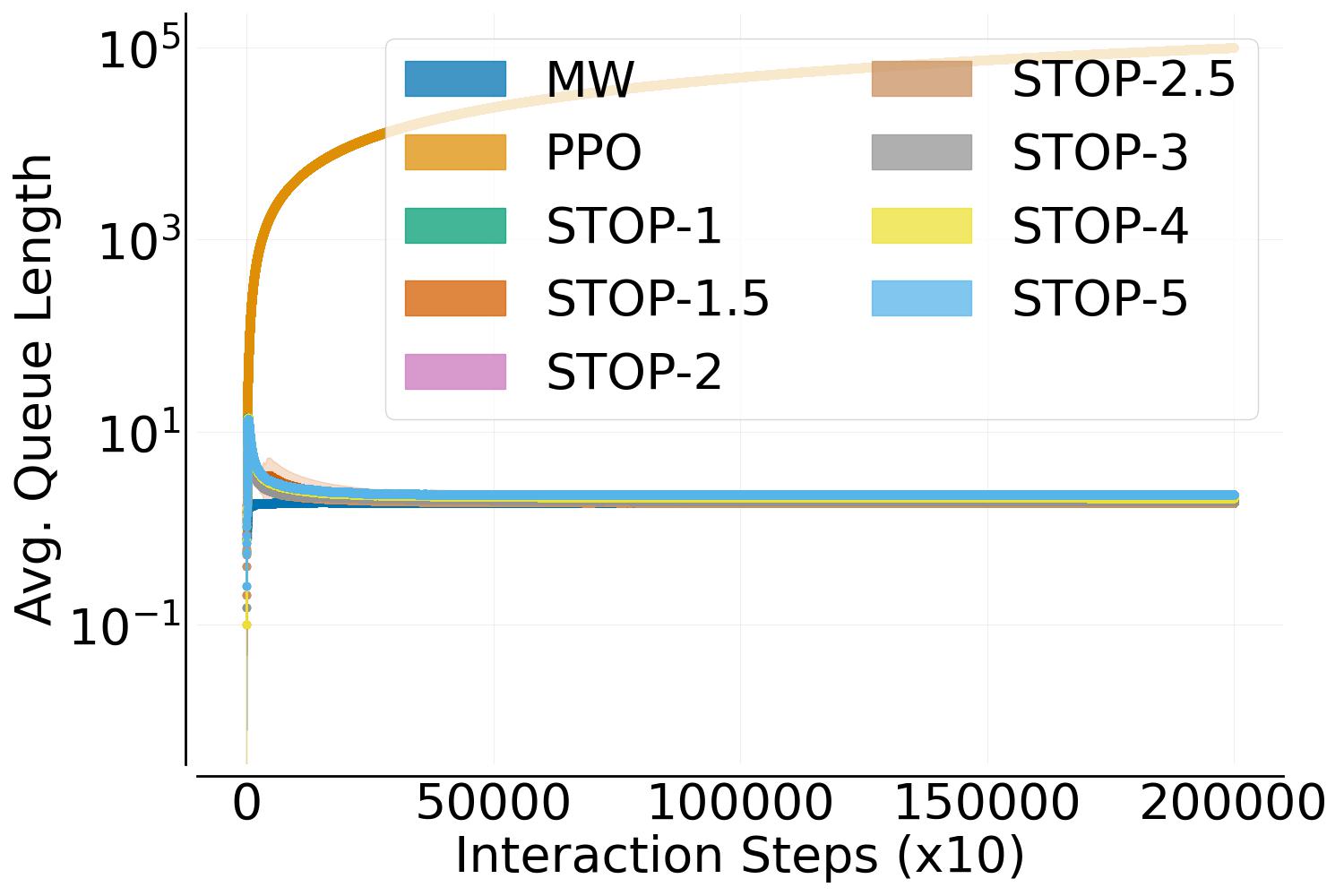}}
        \subfigure[Varying state transformation]{\includegraphics[scale=0.13]{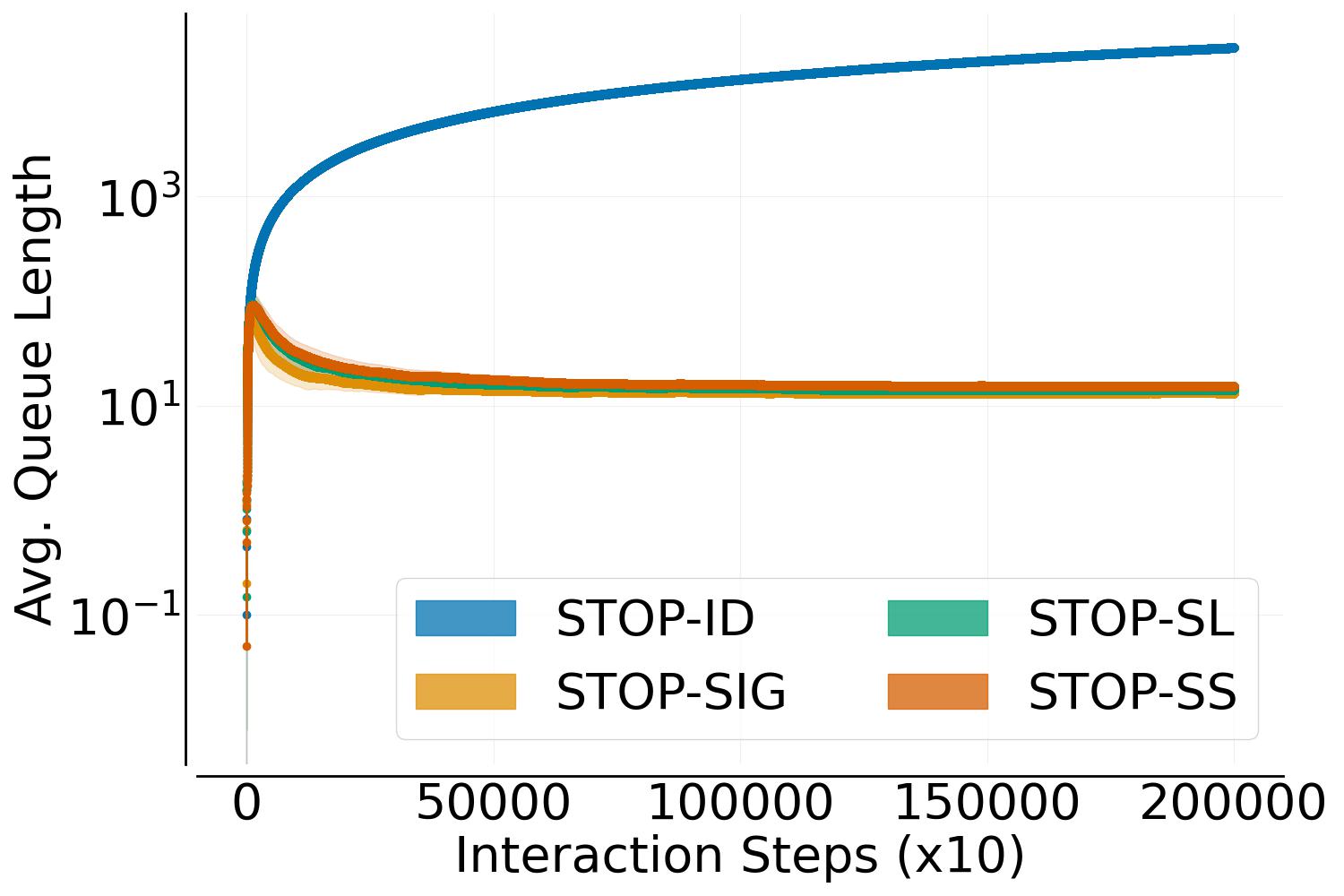}}\\
        \subfigure[High Load]{\includegraphics[scale=0.13]{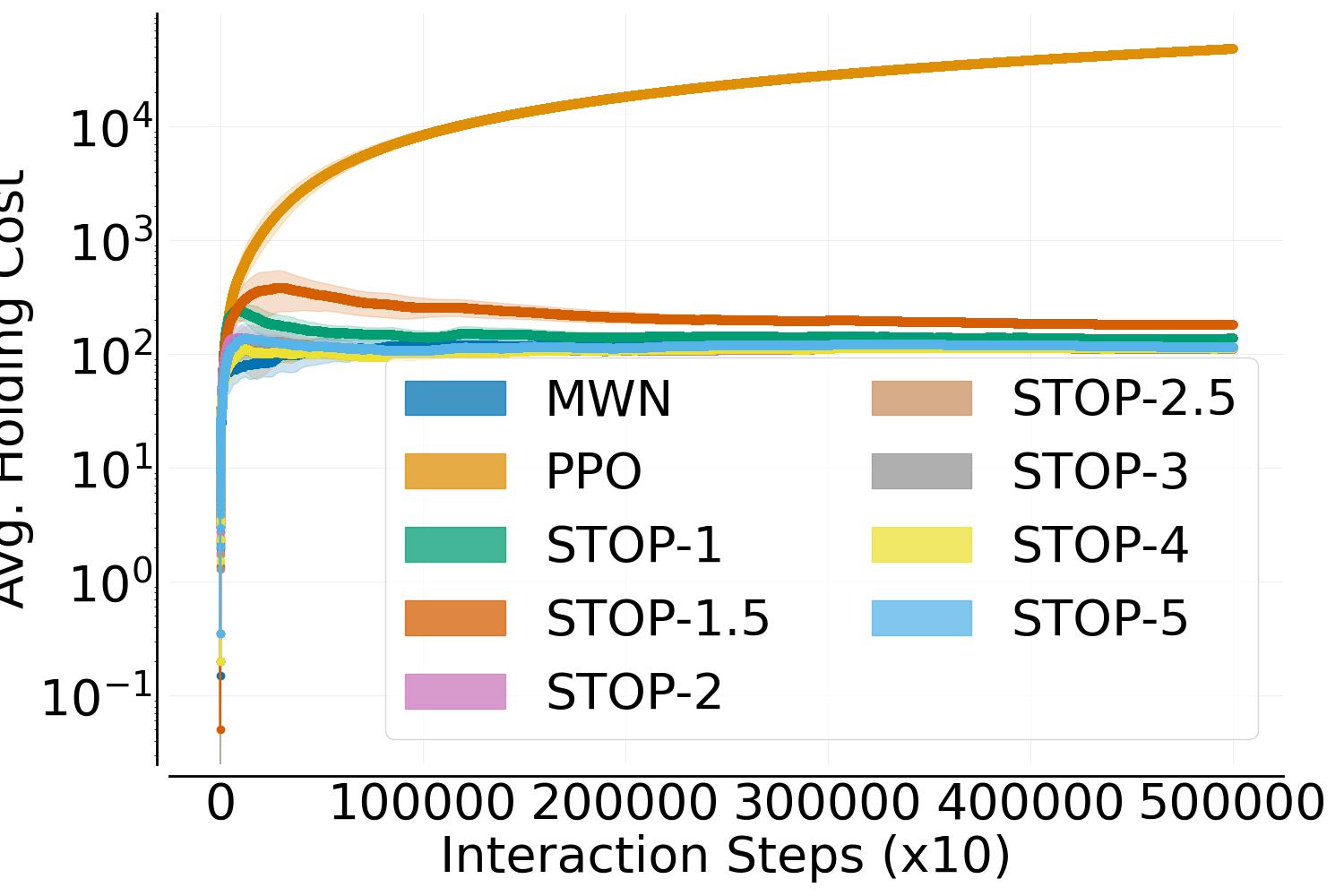}}
        \subfigure[Very High Load - $\#1$]{\includegraphics[scale=0.13]{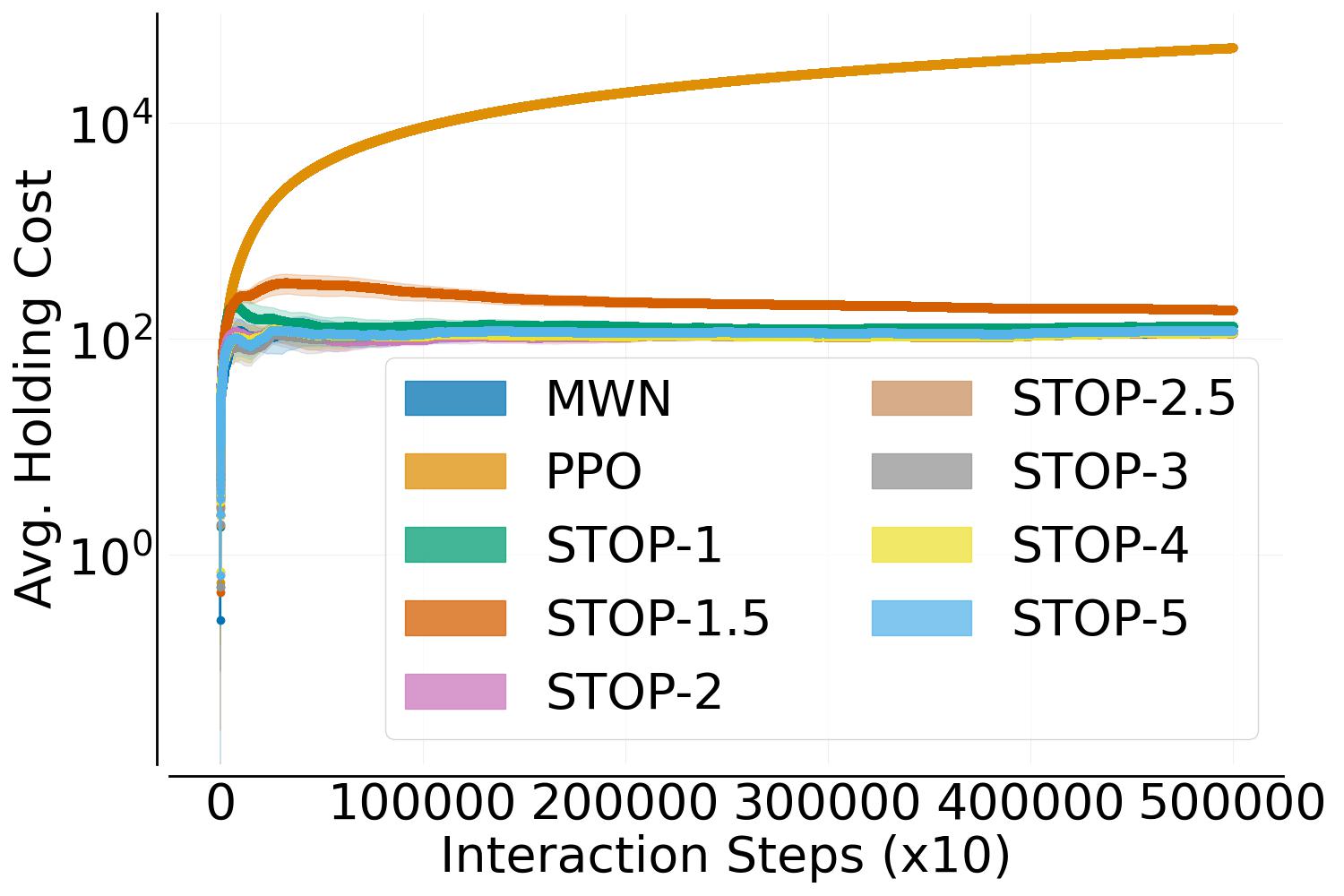}}
        \subfigure[Very High Load - $\#2$]{\includegraphics[scale=0.13]{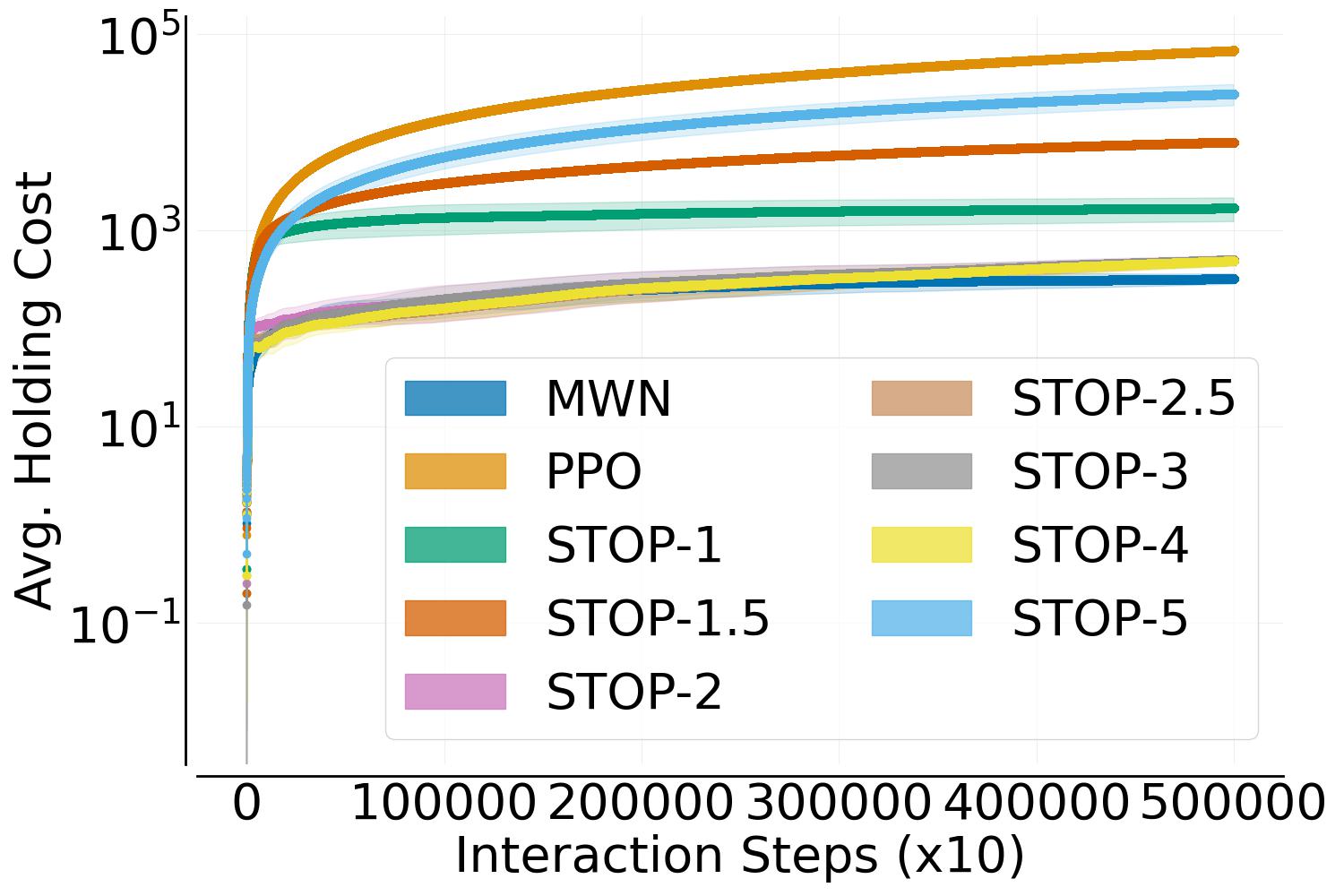}}
    \caption{\footnotesize Zoomed out version of Figure~\ref{fig:base} and Figure~\ref{fig:ablation}. True optimality criterion vs. interaction time-steps on three server-allocation queue networks (top row) and three $N$-network model environments (bottom row). Lower is better. Algorithms are \textsc{ppo} vs. \textsc{stop}-$p$, where $p$ denotes the power of the Lyapunov function. All \textsc{stop} variants use the symloge state transformation.  We also report the performance of \textsc{maxweight} (\textsc{mw}). Recall that unlike \textsc{mw}, \textsc{stop} does not know the transition dynamics. The \textsc{iqm} \cite{agarwal2021deep} is computed of the performance metrics over $20$ trials with $95\%$ confidence intervals. All vertical axes are log-scaled. } 
    \label{fig:base_zoomedout}
\end{figure*}

In this section, we include the remaining main results on a $10$-queue server allocation problem and on the traffic control simulator.  We also include the zoomed-out results from Figure~\ref{fig:base} and Figure~\ref{fig:ablation} in Figure~\ref{fig:base_zoomedout}.
\begin{figure*}[hbtp]
    \centering
        \subfigure[$10$-queues w/o faulty connections (very high load)]
        {\includegraphics[scale=0.11]{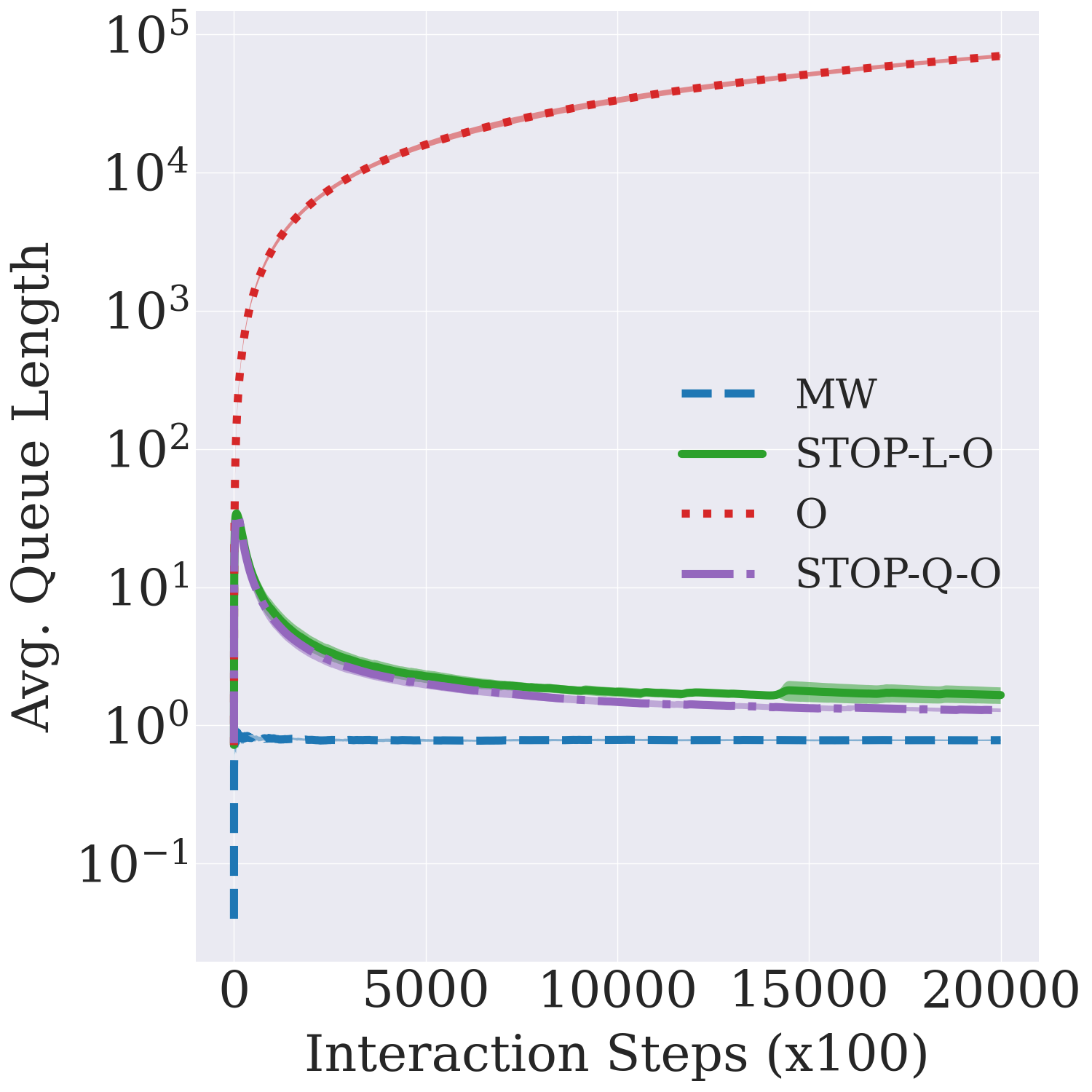}}
        \subfigure[Medium congestion]{\includegraphics[scale=0.11]{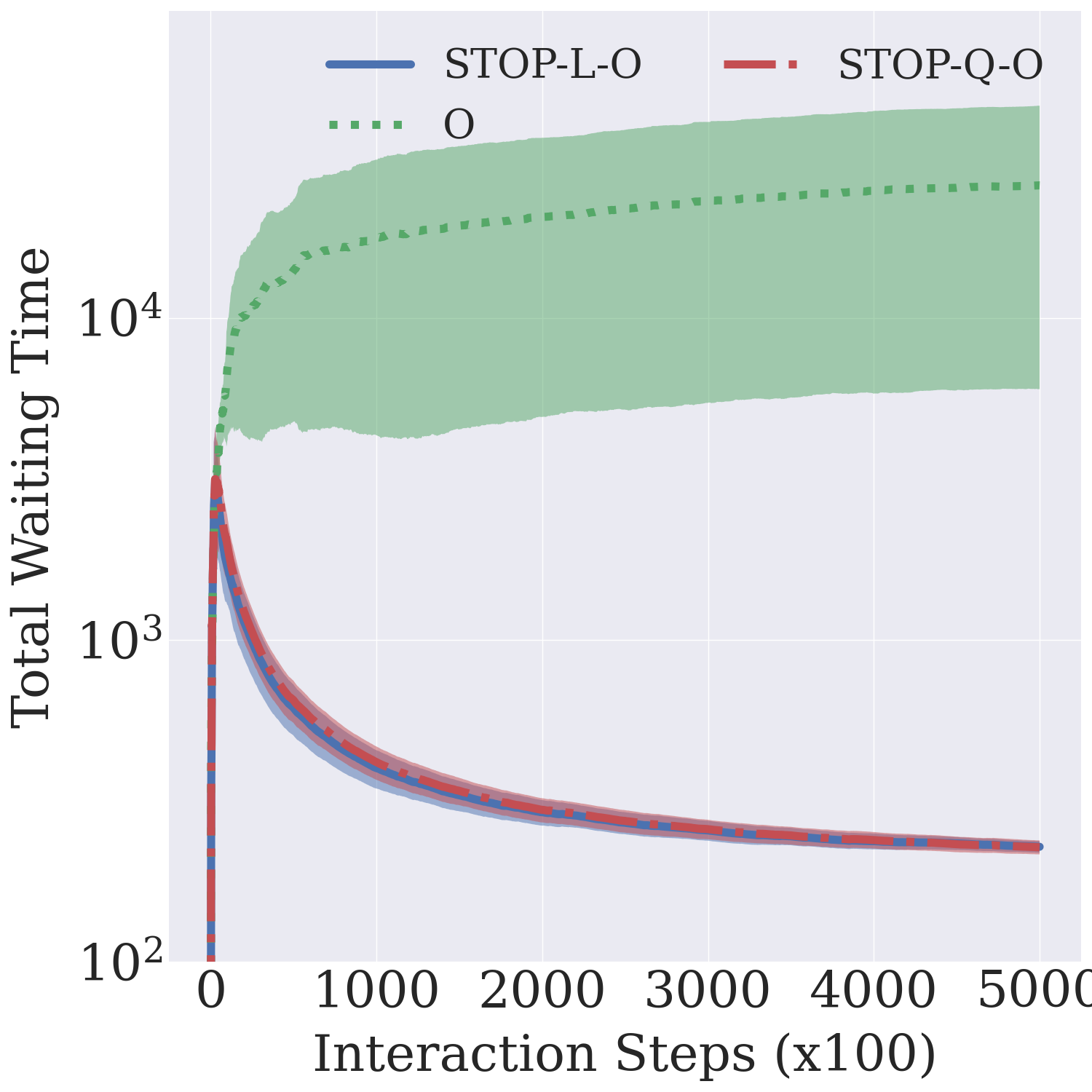}}
        \subfigure[Heavy congestion]{\includegraphics[scale=0.11]{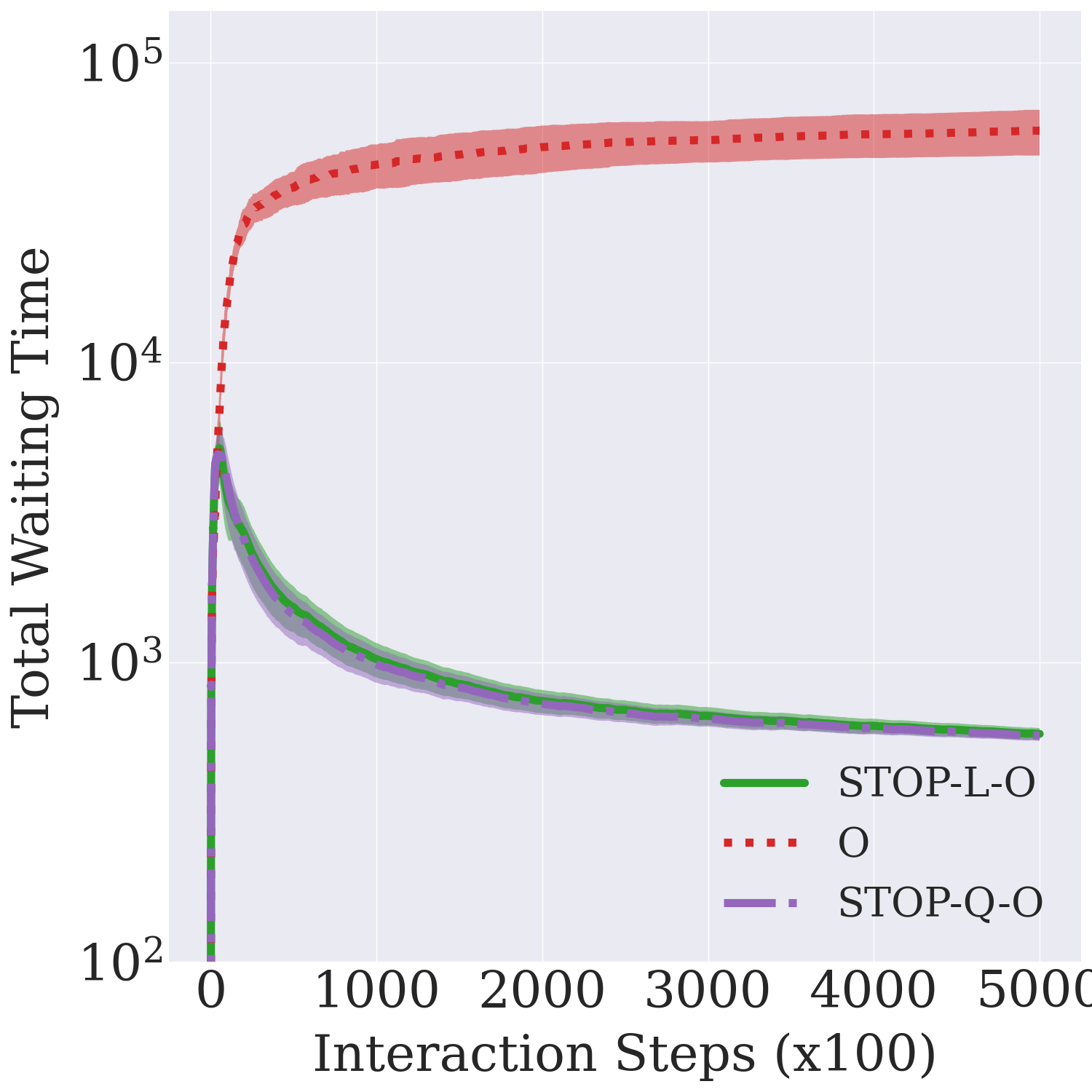}}
        \subfigure[Very heavy congestion]{\includegraphics[scale=0.11]{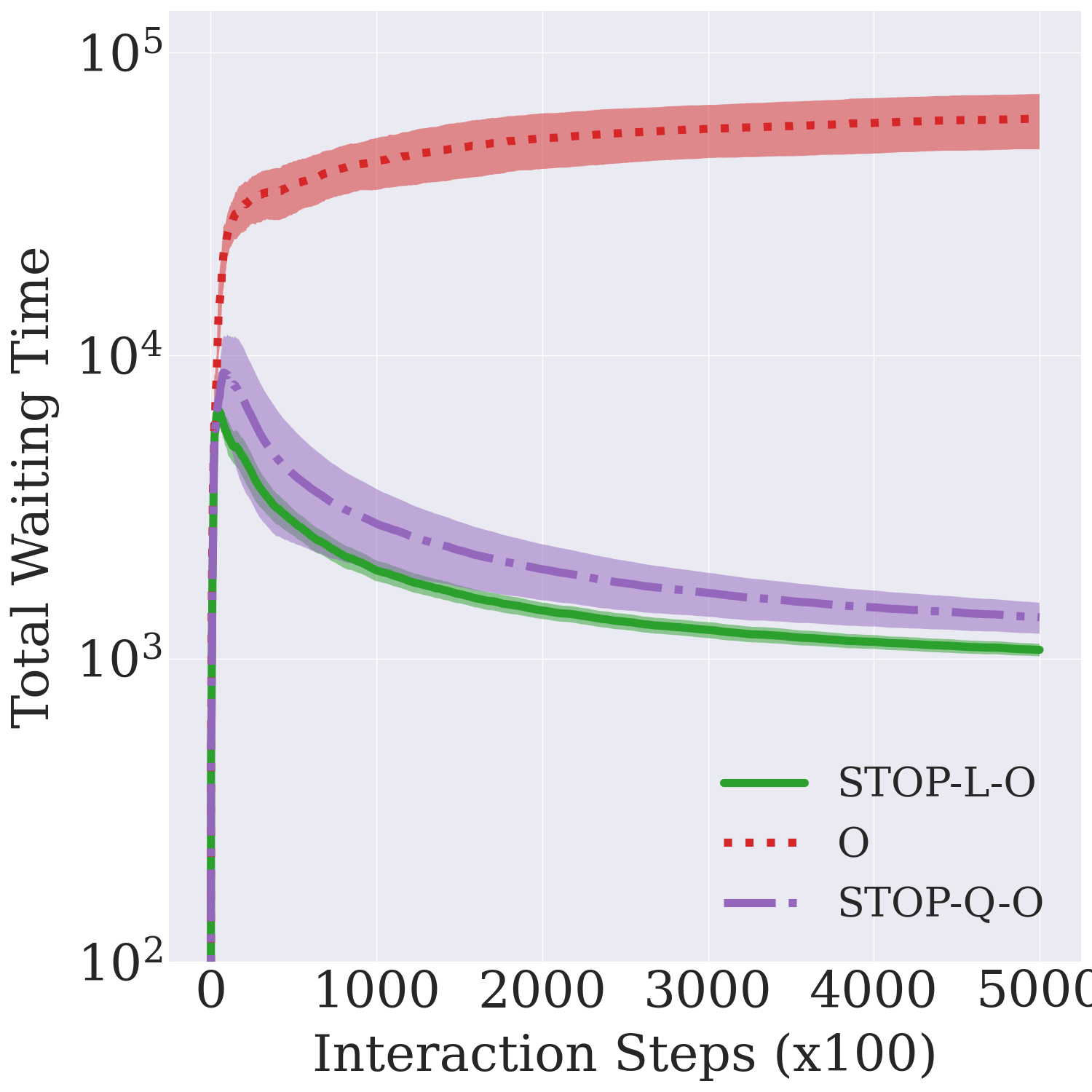}}
    \caption{\footnotesize True optimality criterion vs. interaction time-steps on $10$-queue server allocation (first image) and traffic control environment (remaining three). Lower is better. Algorithms are \textsc{ppo} (\textsc{o}) vs. \textsc{stop} + \textsc{ppo} which use all \textsc{stop} components, where we evaluate the linear (\textsc{stop-l-o}) and quadratic (\textsc{stop-q-o}) stability cost function. For the queuing environment, we also report the performance of \textsc{maxweight} (\textsc{mw}). Recall that unlike \textsc{mw}, \textsc{stop} does not know the transition dynamics. The \textsc{iqm} \cite{agarwal2021deep} is computed of the performance metrics over $15$ trials with $95\%$ confidence intervals.}
    \label{fig:main_res_appendix}
\end{figure*}

\subsection{Environments}
\label{sec:more_env_details}

In this section, we provide additional details about the environments.

\begin{figure*}[hbtp]
    \centering
        \subfigure[Single-server allocation]{\includegraphics[scale=0.3]{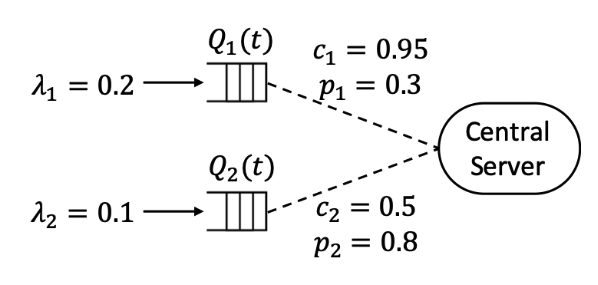}}
        \subfigure[$2$-Model Network]{\includegraphics[scale=0.4]{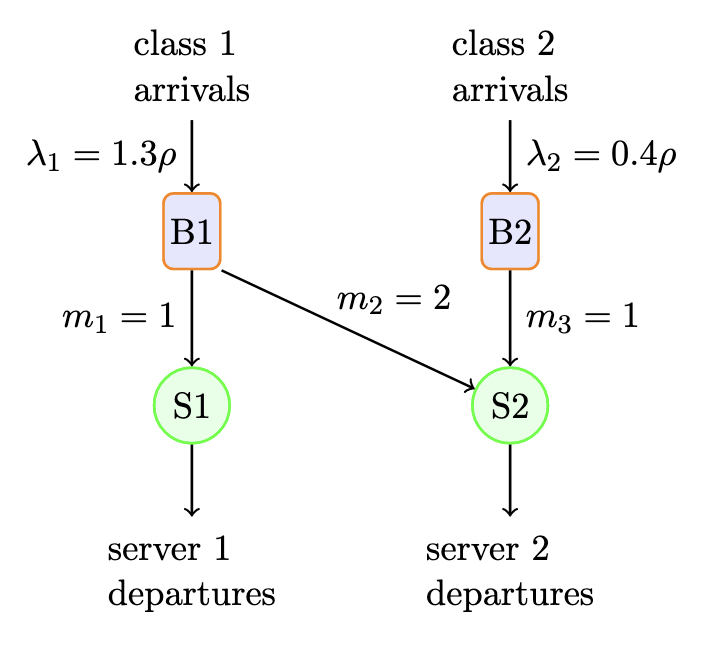}}
        \subfigure[Traffic control]{\includegraphics[scale=0.3]{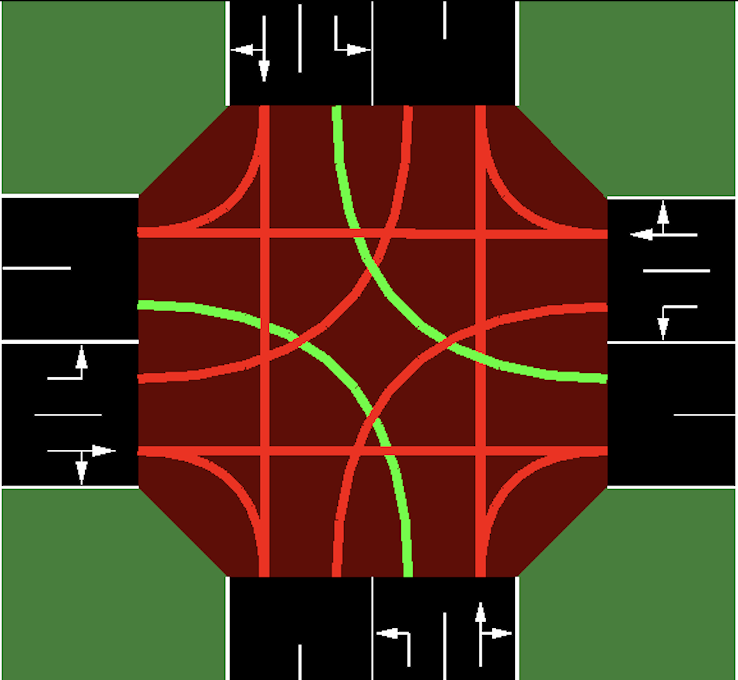}}
    \caption{\footnotesize Left: Server-allocation. Image taken from \citet{liu_rlqueueing_2019}. Center: $2$ model network. Image taken from \citet{dai_ppo_2022}. Right: An example intersection of the traffic control environment. Image taken from \citet{alegre_sumorl_2019}.}
    \label{fig:environments}
\end{figure*}

\paragraph{Single-server allocation queuing} In this environment, there is a single central server that must select among a set of queues to serve. In general, there can be up to $N$ queues (Figure \ref{fig:environments} show a sample $2$ queue setup). At each time-step, new jobs arrive in each queue following a Bernoulli process with probability $\lambda_i$ for queue $i$. Note that at each time-step, at most one job enters each queue, which means more than one job may enter the whole system in total. At each time-step, the server must select among the $N$ queues to serve. A successfully served queue will mean a job will exit that queue, so at most one job can exit the system at a given time-step. When a server selects a queue $i$, the server succeeds in making a job exit only if: it can connect to queue $i$ (which is dependent on the connectivity probability, $c_i$) and the job is successfully served (which is dependent on the service probability, $p_i$). The state of the server is queue length of each queue and $0/1$ flag indicating whether the server can connect to a specific queue, resulting in $2N$-dimensional state. The action space is index of the queue, resulting in $N$ dimensions. The goal is to minimize the average queue length. In the non-faulty connection setting, all the connectivity flags are $1$. The optimal policy in the faulty connection setting is an open problem \citep{ganti_qopen-prob_2007}. We consider settings where the system is stabilizable i.e. $\sum_i^N \frac{\lambda_i}{p_i} < 1 - \prod_i^N (1-c_i)$ and $\frac{\lambda_i}{p_i} < c_i$.

The Bernoulli probability parameters of the tested environments are:
\begin{enumerate}
    \item $2$-queue \emph{without} faulty connections (see Figure~\ref{fig:2queue_example}) (medium load)
    \begin{itemize}
        \item Arrival rates: $\lambda_1 = 0.2$, $\lambda_2 = 0.1$
        \item Service rates: $p_1 = 0.3, p_2 = 0.8$
        \item Connection probabilities: $c_1 = 1, c_2 = 1$
    \end{itemize}
    \item $2$-queue with faulty connections (high load)
    \begin{itemize}
        \item Arrival rates: $\lambda_1 = 0.2$, $\lambda_2 = 0.1$
        \item Service rates: $p_1 = 0.3, p_2 = 0.8$
        \item Connection probabilities: $c_1 = 0.95, c_2 = 0.5$
    \end{itemize}
    \item $2$-queue with faulty connections (very high load)
    \begin{itemize}
        \item Arrival rates: $\lambda_1 = 0.2$, $\lambda_2 = 0.1$
        \item Service rates: $p_1 = 0.3, p_2 = 0.8$
        \item Connection probabilities: $c_1 = 0.7, c_2 = 0.5$
    \end{itemize}
    \item $10$-queue with non-faulty connections (very high load)
    \begin{itemize}
        \item Arrival rates: $\lambda_1 = 0.05, \lambda_2 = 0.01, \lambda_3 = 0.2, \lambda_4 = 0.4, \lambda_5 = 0.05, \lambda_6 = 0.01, \lambda_7 = 0.02, \lambda_8 = 0.01, \lambda_9 = 0.015, \lambda_{10} = 0.01$
        \item Service rates: $p_1 = 0.9, p_2 = 0.85, p_3 = 0.95, p_4 = 0.75, p_5 = 0.9, p_6 = 0.9, p_7 = 0.85, p_8 = 0.9, p_9 = 0.9, p_{10} = 0.85$
        \item Connection probabilities: $c_i = 1$ for all $1\leq i\leq 10$
    \end{itemize}
\end{enumerate}

When evaluating the \textsc{rl} algorithms, we compare their performance to \textsc{maxweight} \citep{tassiulas_mw_1990, stolyar_mw_2004}, a well-known algorithm that achieves stability for a certain class of queuing scenarios, but which relies on the knowledge of the system model (i.e., some parts of the transition dynamics) and it is generally unknown how far \textsc{maxweight} is from optimality. It is a very strong non-RL baseline from decades of research from the stochastic networking community. 

\paragraph{$2$-model Network}
In this domain, there are two queues of jobs for class 1 and class 2 jobs, $B_1$ and $B_2$. These jobs come into the queues following a Poisson process with arrival rates $\lambda_1$ and $\lambda_2$, which is a function of $\rho$ which determines the load. Class 1 jobs can be processed by server S1 as well as server S2, while class 2 jobs can only be processed by server S2. The success rate of S2 serving class 1 jobs is given by the service rate $\mu_2 = 1/m_2$ and of serving class 2 jobs is $\mu_3 = 1/m_3$. Similarly, success of S1 serving class 1 jobs is given by $\mu_1 = 1/m_1$. All these success rates are exponentially distributed with mean $m_i$. The holding cost of keeping a job waiting in $B_1$ is $3$ and in $B_2$ is $1$. The agent in this case is S2, which must decide whether to serve class 1 jobs or class 2 jobs. Its state is the queue lengths of $B_1$ and $B_2$. Its action is the discrete action denoting the index of the selected queue. The optimality criterion is the average holding cost per time-step i.e. $3x_1 + x_2$, where $x_i$ is the number of waiting jobs in queue $i$. For more details, refer to \citet{dai_ppo_2022}. We refer to their code for the environment: \url{https://github.com/mark-gluzman/NmodelPPO/blob/master/NmodelDynamics.py}. We consider settings where i.e. $\frac{\lambda_1 - \mu_1}{\mu_2} \leq 1 - \frac{\lambda_2}{\mu_3}$.

The parameters of the environments we evaluated on are:
\begin{enumerate}
    \item High load
    \begin{itemize}
        \item $\rho = 0.99$ and using parameters exactly as in Figure~\ref{fig:environments}.
    \end{itemize}
    \item Very high load $\#1$
    \begin{itemize}
        \item $\rho = 0.99$ and using parameters exactly as in Figure~\ref{fig:environments}.
    \end{itemize}
    \item Very high load $\#2$
    \begin{itemize}
        \item $\rho = 0.95$
        \item Arrival rates: $\lambda_1 = 0.9$, $\lambda_2 = 0.8$
        \item Service rates: $\mu_1 = 1, \mu_2 = 0.9, \mu_3 = 0.8$
    \end{itemize}
\end{enumerate}

\paragraph{Total Queue Length as Optimality Objective}. While the lessons in our paper are generally applicable to \textsc{rl}, our work is grounded in queuing theory. As such, we are specifically interested in learning control policies that minimize and bound the \emph{system latency}. Therefore, according to Little's Law~\citep{garcia_little_2008}, we seek to minimize the true optimality cost function, the total queue length, $c(s, a, s')=\|s'\|_1$, where states $s$ and $s'$ are vectors consisting of the current and next queue lengths of each queue in the system and $a$ is the action taken.

\paragraph{Traffic control} 

In this environment, a traffic controller must select from a set of phases (shown in green in Figure \ref{fig:environments}), a set of non-conflicting lanes, to allow cars to move. At each time-step, new cars arrive in each lane at different rates, which determines the traffic congestion level. In our experiments, we considered medium to very high levels of traffic congestion. The state is the number of cars waiting in each lane along with indicator flags for which lanes have a green and yellow light. The action space is the number of phases. The state space is $21$ dimensions and the action space is $4$. The goal is to minimize the total waiting time of all the cars. To model a real-life traffic situation, the \textsc{sumo} simulator places a cap of $\approx 100$ on each lane. We use the \textsc{sumo} simulator implementation \citep{behrisch_sumo_2011,alegre_sumorl_2019}.

For exact traffic demands used in the experiments, see the \texttt{sumo/nets/big-intersection/generator.py} file in the attached code.

\subsection{Additional Empirical Setup Details}

\paragraph{PPO Training} We train average-reward \textsc{ppo} \citep{zhang_artrpo_2021, dai_ppo_2022} using the default hyperparameters (network architecture, learning rate, mini batches, epochs over the dataset etc.) in the cleanRL code base \citep{huang2022cleanrl}. For all algorithms and variations, we set the interval between policy updates during the interaction (rollout buffer length) to be $200$ . The agent starts in a randomly initialized state, takes a number of actions until it fills up its rollout buffer and then makes policy updates using this buffer, and the process repeats. We normalize the advantages in the rollout buffer by dividing each by the standard deviation computed over the buffer. As suggested by \citet{dohare2023overcoming}, we set Adam's beta values to be $\beta_1 = \beta_2 = 0.9$.

\subsection{Hardware For Experiments}
For all experiments, we used the following compute infrastructure:
\begin{itemize}
    \item Distributed cluster on HTCondor framework
    \item Intel(R) Xeon(R) CPU E5-2470 0 @ 2.30GHz
    \item RAM: 5GB
    \item Disk space: 5GB 
\end{itemize}

\begin{figure*}[hbtp]
    \centering
        \subfigure[Other bounded formulations of true cost]{\includegraphics[scale=0.15]{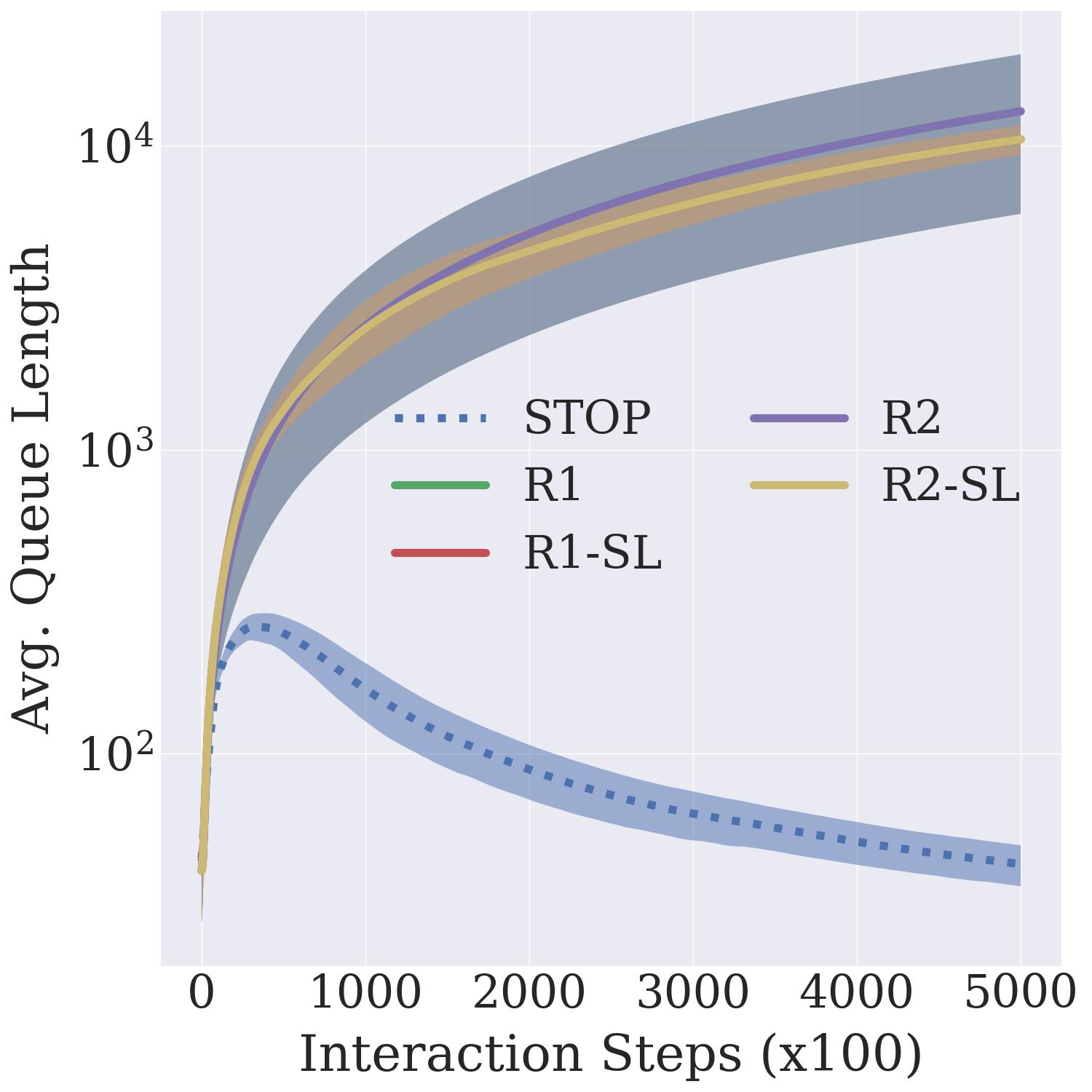}\label{fig:other_reward_formulations}}
        \subfigure[Training wheels comparison]{\includegraphics[scale=0.15]{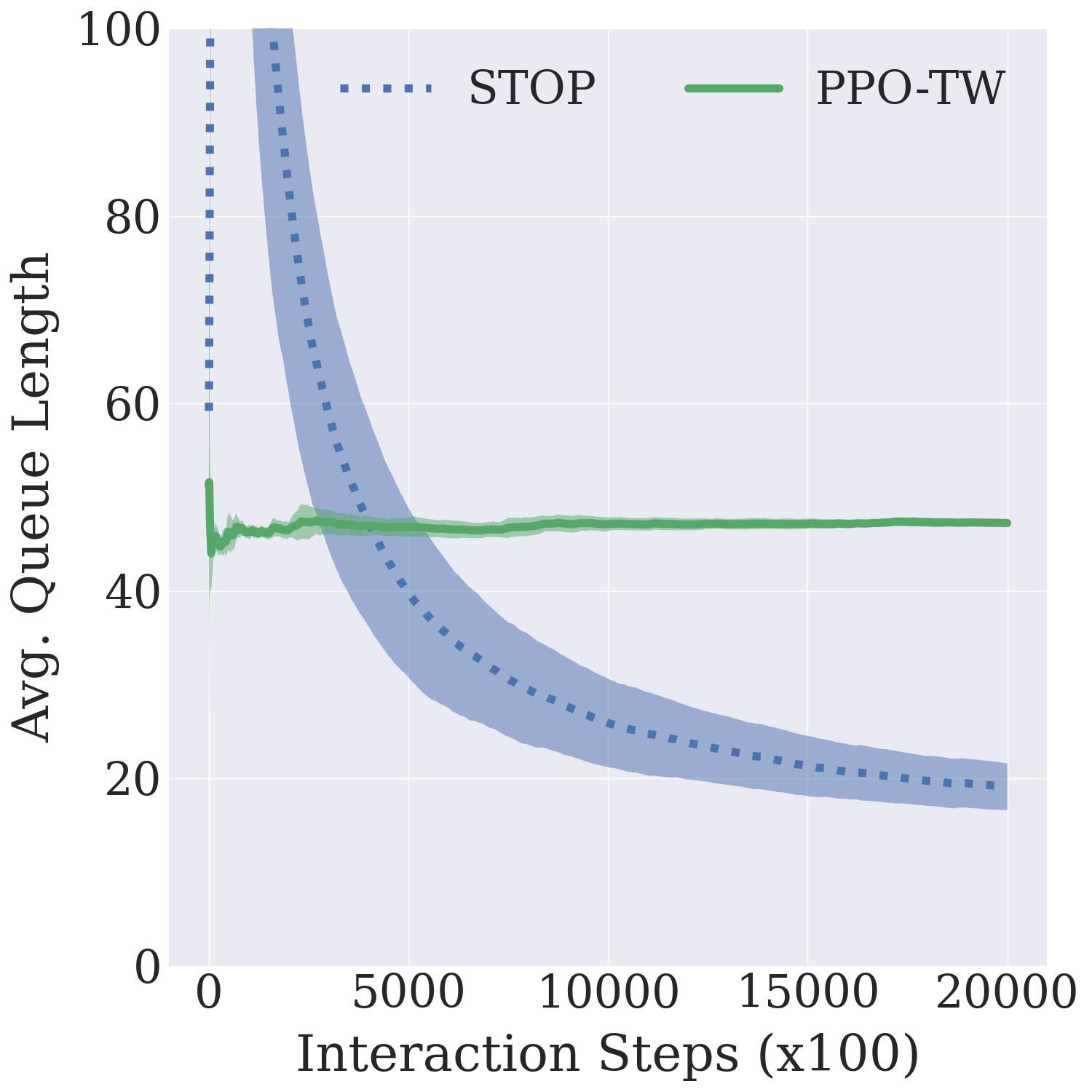}\label{fig:rl_training_wheels}}
    \caption{\footnotesize Average queue length on the $2$-queue problem w/ faulty connections over time by (a) different \textsc{rl}  agents optimizing different cost formulations vs. our \textsc{stop} agent and (b) \textsc{ppo-tw}. Performance metrics are computed over $5$ trials with $95\%$ confidence intervals. Lower is better.}
    
\end{figure*}

\section{Transforming Optimality Cost Function Leads to Instability}

In Figure~\ref{fig:other_reward_formulations} we also include the performance of \textsc{rl} agents optimizing other transformations of the true optimality cost function $c(s_{t+1})$. We consider the transformed cost: 1) $c'(s_t,a_t, s_{t+1}) := -\text{exp}(-||s_{t} + a_t||^2_2)$ and 2) $c'(s_t,a_t, s_{t+1}) := -\text{exp}(-||s_{t+1}||^2_2)$. The former is $R_1$ and latter is $R_2$ in the plot. We show performance  when these agents do not use any state transformation and when they use the symloge transformation ($-SL$). We find that the agents still unstable, thus providing further evidence that simply transforming the true cost function in this class of problems is insufficient to yield good performance.

\section{PPO With Training Wheels}
\label{sec:ppo_tw}

In this section, we provide preliminary evidence that equipping \textsc{ppo} with training wheels i.e. a stable policy may perform worse than \textsc{stop}.

In this experiment, we evaluate \textsc{ppo} with training wheels (\textsc{ppo-tw}). The \textsc{ppo-tw} setup closely models that of \cite{mao2019TowardsSO} where we equip an on-policy policy gradient algorithm (\textsc{ppo}) with a stable policy. In our case, the stable policy is \textsc{maxweight} \citep{stolyar_mw_2004, tassiulas_mw_1990}.  \textsc{maxweight} is deployed if the maximum queue length of the system exceeds $100$, at which point \textsc{maxweight} is used until it drives the system's maximum queue lengths to less than $50$. Once it has done that, the \textsc{ppo} policy is deployed. Note that 1) \textsc{maxweight}  relies on knowing information of the transition dynamics, which can be limiting. \textsc{stop}, on the other hand, does not assume access to such knowledge and 2) \textsc{ppo-tw} optimizes the true optimality cost (average queue length).

From Figure~\ref{fig:rl_training_wheels}, we find that while \textsc{stop} performs poorly during initial phases of learning, it is able to significantly outperform \textsc{ppo-tw} later on. \textsc{stop} is able to learn the stabilizing and optimal actions from a destabilizing, random policy. In the case of \textsc{ppo-tw}, however, since the initial \textsc{rl} policy is poor (random) it causes the agent to diverge, which violates the safety condition often, which results in frequent deployment of the stable policy. However, this off-policy data cannot be used to update the \textsc{ppo} policy. Thus, the \textsc{ppo} policy continues to remain poor since it has inadequate data to train on, which causes the agent to diverge until it violates the safety condition, at which point the stable policy is deployed again. As noted by \cite{mao2019TowardsSO}, the off-policy data generated by the stable policy cannot be used to train the \textsc{ppo} policy. As we have noted in our future work as well, an interesting further direction will be to apply our ideas to off-policy algorithms.

\end{document}